\documentclass[journal]{IEEEtran}
\usepackage{booktabs}
\usepackage{subfigure}
\usepackage{graphicx}
\usepackage{float}
\usepackage{verbatim}
\usepackage{graphicx,amssymb,mathrsfs,amsmath,latexsym}
\usepackage{epsfig}
\usepackage{booktabs}
\usepackage{epstopdf}
\usepackage{color,xcolor,url}
\usepackage{algpseudocode}
\usepackage[linesnumbered,ruled]{algorithm2e}
\usepackage{cite}
\usepackage{rotating}
\usepackage[colorlinks,urlcolor=blue,driverfallback=dvipdfm]{hyperref}
\usepackage{multirow}
\usepackage{array}

\newcolumntype{R}[1]{>{\PreserveBackslash\raggedleft}p{#1}}
\newcolumntype{L}[1]{>{\PreserveBackslash\raggedright}p{#1}}

\makeatletter
\def\hlinew#1{%
  \noalign{\ifnum0=`}\fi\hrule \@height #1 \futurelet
   \reserved@a\@xhline}

\newcommand{\PreserveBackslash}[1]{\let\temp=\\#1\let\\=\temp}

\begin{document}

\title{Deep Learning for UAV-based Object Detection and Tracking: A Survey}

\author{Xin~Wu,~\IEEEmembership{Member,~IEEE,}
        Wei~Li,~\IEEEmembership{Senior Member,~IEEE,}
        Danfeng~Hong,~\IEEEmembership{Senior Member,~IEEE,}
        Ran Tao,~\IEEEmembership{Senior Member,~IEEE,}
        and~Qian~Du,~\IEEEmembership{Fellow,~IEEE}
        
\thanks{This work was supported, in part by the National Natural Science Foundation of China under Grant 61922013, 62101045, and U1833203, and partly by the China Postdoctoral Science Foundation Funded Project No. 2021M690385.} 
\thanks{X. Wu, W. Li, and R. Tao are with the School of Information and Electronics, Beijing Institute of Technology, 100081 Beijing, China, and Beijing Key Laboratory of Fractional Signals and Systems, 100081 Beijing, China. (e-mail: 040251522wuxin@163.com; liwei089@ieee.org; rantao@bit.edu.cn)}
\thanks{D. Hong is with the Key Laboratory of Digital Earth Science, Aerospace Information Research Institute, Chinese Academy of Sciences, 100094 Beijing, China. (e-mail: hongdf@aircas.ac.cn)}
\thanks{Q. Du is with the Department of Electrical and Computer Engineering, Mississippi State University, Starkville, MS 39762, USA. (e-mail: du@ece.msstate.edu)}
}

\markboth{IEEE GRSM 2021, XX, XX}%
{Shell \MakeLowercase{\textit{et al.}}: Bare Demo of IEEEtran.cls for IEEE Journals}

\maketitle

\begin{abstract}
\textcolor{blue}{This is the pre-acceptance version, to read the final version please go to IEEE Geoscience and Remote Sensing Magazine on IEEE Xplore.} Owing to effective and flexible data acquisition, unmanned aerial vehicle (UAV) has recently become a hotspot across the fields of computer vision (CV) and remote sensing (RS). Inspired by recent success of deep learning (DL), many advanced object detection and tracking approaches have been widely applied to various UAV-related tasks, such as environmental monitoring, precision agriculture, traffic management. This paper provides a comprehensive survey on the research progress and prospects of DL-based UAV object detection and tracking methods. More specifically, we first outline the challenges, statistics of existing methods, and provide solutions from the perspectives of DL-based models in three research topics: object detection from the image, object detection from the video, and object tracking from the video. Open datasets related to UAV-dominated object detection and tracking are exhausted, and four benchmark datasets are employed for performance evaluation using some state-of-the-art methods. Finally, prospects and considerations for the future work are discussed and summarized. It is expected that this survey can facilitate those researchers who come from remote sensing field with an overview of DL-based UAV object detection and tracking methods, along with some thoughts on their further developments.
\end{abstract}

\begin{IEEEkeywords}
Deep learning, object detection, object tracking, remote sensing, unmanned aerial vehicle, video.  
\end{IEEEkeywords}

\section{Introduction}
\IEEEPARstart{O}{bject} detection and tracking, as an important research topic in the field of remote sensing, has been widely investigated and applied to various civil and military tasks, such as environmental monitoring, geological hazard detection, precision agriculture, and urban planning. Traditional object acquisition methods derive mainly from satellites and manned aircraft. Normally, the two types of platforms run on a fixed rail or follow a predetermined path, or temporarily change the running route and hover according to a commissioned task, e.g., city planning and mapping, or performing object observation in a harsh and inhospitable environment, e.g., remote sensing in the cryosphere. However, the cost of satellites and manned aircraft, and the potential safety issues of pilots inevitably limit the application scope of such platforms.
\begin{figure}[!t]
    \centering\includegraphics[width=0.95\linewidth]{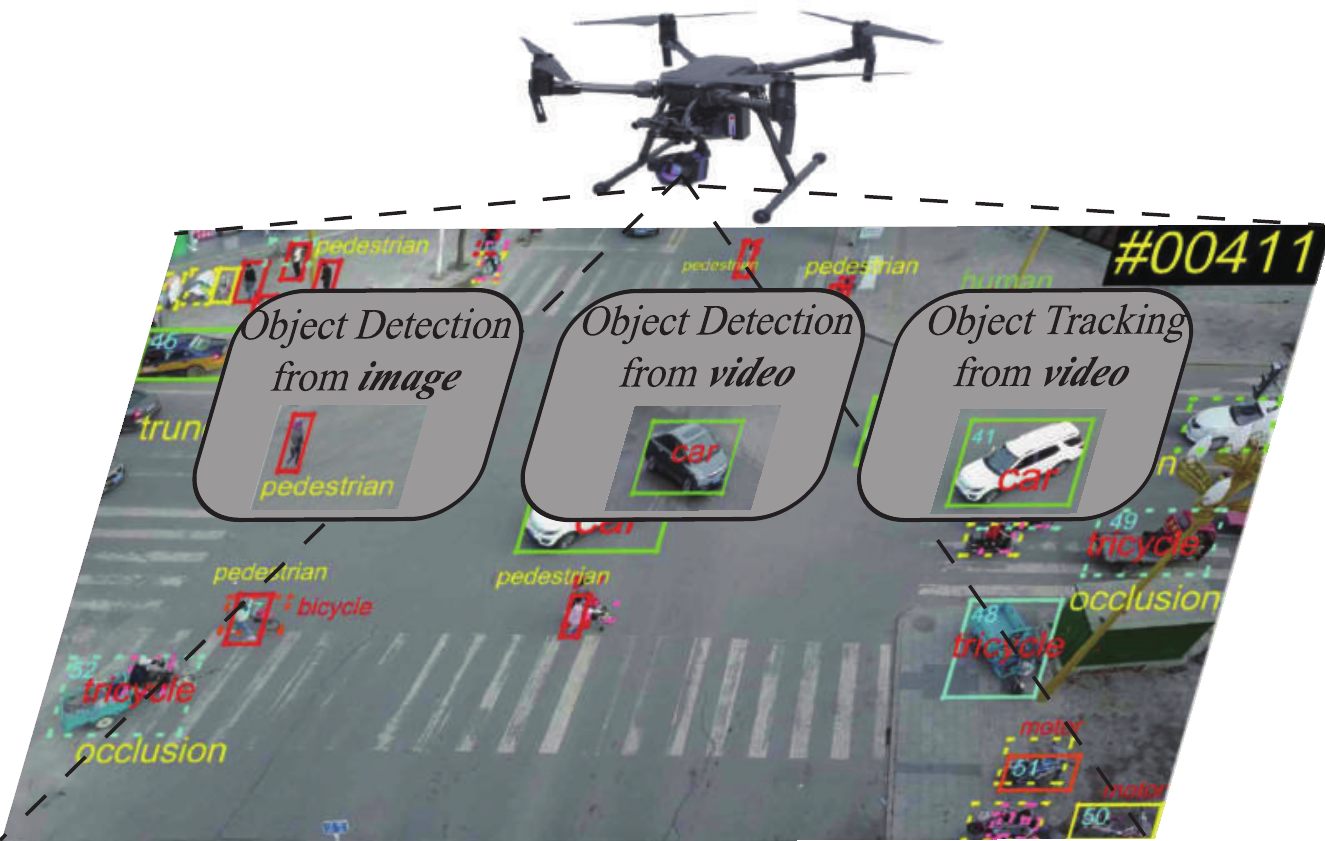}
    \caption{A complex urban scenario for UAV object detection and tracking. For simplicity, only bounding boxes and class names for certain objects are drawn in the imagery.}
\label{fig:outline}
\end{figure}

With the development of microelectronic software and hardware, navigation and communication technology renewal, and breakthroughs in materials and energy technology, unmanned aerial vehicle (UAV) platform already an international research hotspot in remote sensing has rapidly emerged. A UAV remote sensing system is a high-tech combination of science and technology integrated UAVs, remote sensing, global positioning system (GPS) positioning, and inertial measurement unit (IMU) attitude determination means. It is a dedicated remote sensing system with the goal of obtaining low-altitude high-resolution remote sensing images. Compared with traditional platforms, UAV makes up for information loss caused by weather, time, and other limitations. In addition, the high mobility of UAVs enables it to flexibly collect video data without geographic restriction. These data, either in contents or time, are extremely informative, and thus object detection and tracking have entered the era of mass UAV \cite{nrm1,8682048,GISReview}, which has played an increasingly important role in land cover mapping \cite{hong2019learnable,hong2021more}, smart agriculture \cite{ar1,ar2}, smart city \cite{menouar2017uav}, traffic monitoring \cite{honkavaara2013processing}, and disaster monitoring \cite{erdelj2017help}, among other topics. 

As one of the fundamental computer vision problems, object detection and tracking employ classic, i.e., statistically-based, methods \cite{li2004statistical,haag1999combination}. However, today's massive quantities of data impact the performance of these traditional methods, which poses a problem for feature dimension explosion, yielding higher storage space and time costs. Owing to the emergence of deep neural network (DL) techniques \cite{rasti2020feature,wu2020vehicle,hong2021graph}, hierarchical feature representations with enough sample data can be learned with deep and complex networks. Since 2015, the deep neural network has become a mainstream framework used for UAV object detection and tracking \cite{liu2020uav,cai2019guided}. Fig. \ref{fig:outline} shows an example of object detection and tracking in an urban areas with UAV remote sensing. Classic deep neural networks are divided into two major categories: two-stage and one-stage networks. Among them, the two-stage networks, such as RCNN \cite{girshick2014rich}, Fast RCNN \cite{girshick2015fast}, and Faster RCNN \cite{ren2015faster}, first need to generate a region proposal (RP), and then classify and locate candidate regions. A series of work \cite{zhu2018vision, deng2018multi, wu2018msri} has demonstrated that a two-stage network is suitable for applications with higher detection accuracy. A one-stage network, such as SSD \cite{liu2016ssd} and YOLO \cite{liu2020uav,luo2020fast,redmon2016you}, directly generates class probability and coordinate position, and is faster than a two-stage network. Similarly, there are some faster light weight networks, such as mobilenet SSD \cite{howard2017mobilenets}, YOLOv3 \cite{redmon2018yolov3}, ESPnet\_v2 \cite{mehta2019espnetv2}, etc. Therefore, one-stage and faster light weight networks are the final winners for UAV remote sensing practical applications with high-speed requirements. But for low-resolution data, it fails to produce good results without preprocessing images or modifying the classic neural network structure. 
        
This paper focuses on UAV with a maximum take-off weight of fewer than 30 kilograms, and provides a comprehensive review of deep learning (DL)-based UAV object detection and tracking methods by summarizing the latest published work, discussing the key issues and difficult problems, and delineating areas of future development. 

\begin{figure}[!t]
    \centering
    \subfigure{
             \includegraphics[width=0.45\textwidth]{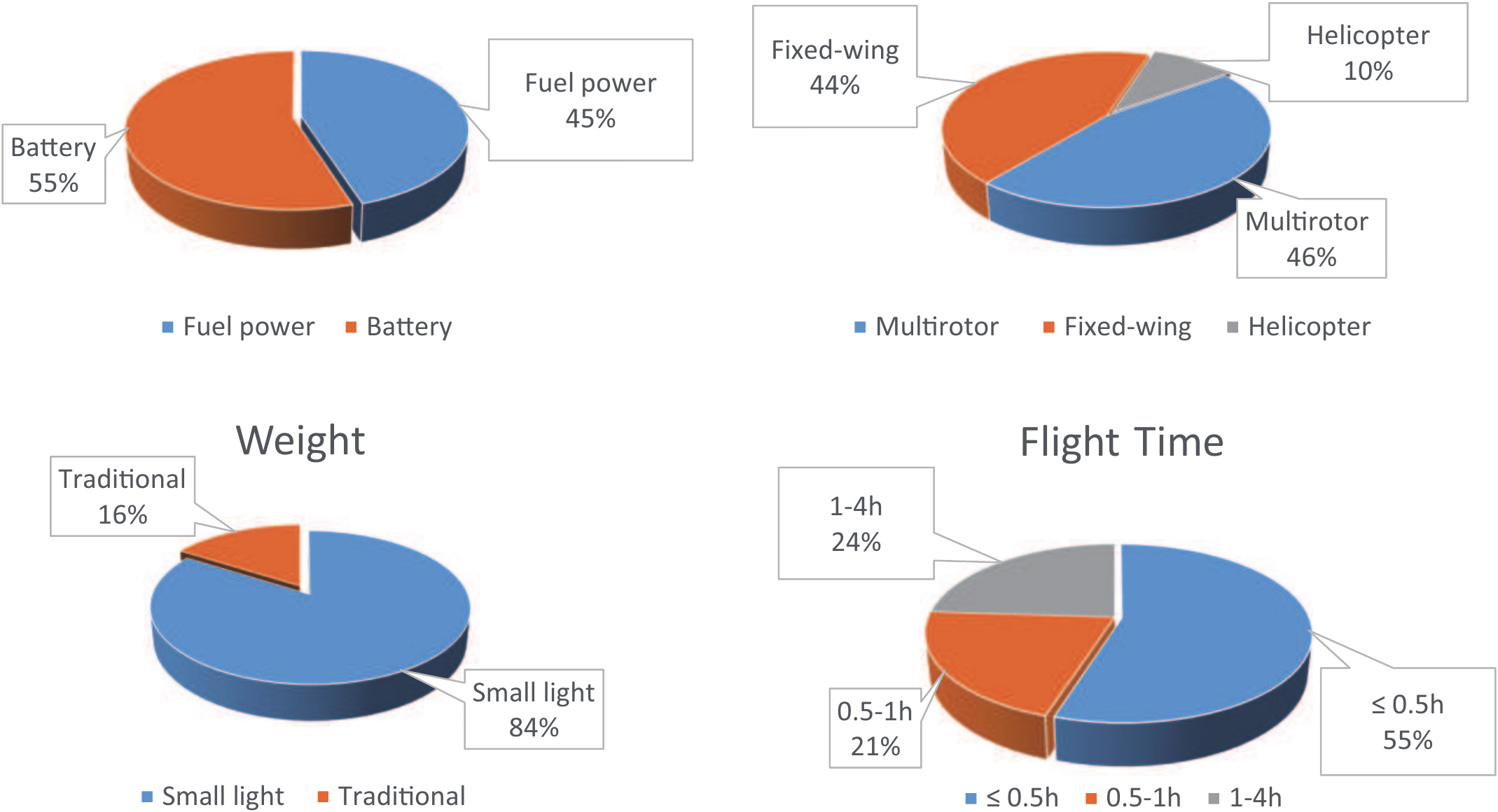}
    }
    \caption{Partial statistical analysis results of light and small UAVs currently in use.}
\label{fig:UAV_Statistic}
\end{figure}

The remainder of this paper is organized as follows. Section \ref{sec:Re} briefly summarizes the statistics of UAV aircraft and related publications. Section \ref{sec:Data} describes the existing UAV-based remote sensing datasets. Section \ref{sec:VII}-\ref{sec:MOT} reviews the existing DL-based work closely related to UAV-based object detection and tracking for the three sub-branches. Section \ref{sec:Con} discusses conclusions. 

\section{Related Surveys and Brief Statistics} \label{sec:Re}
\subsection{UAV Aircraft Statistics} \label{subsec:UAV aircraft}
Fig. \ref{fig:UAV_Statistic} shows the classification of UAVs in present use through statistical analysis. From the perspective of power supply, battery power is used more often than fuel power; for the aerodynamic shape, multi-rotor is more common than fixed-wing; for the weight of the aircraft, the majority is under 30 kilograms, which is considered small light UAV; and the flight time for most UAVs is less than 1 hour. The quantitative analysis results show that small light UAVs have become the main type used for study and application, and have more market weight. In addition, the ``Small light UAV remote sensing development report'' published in 2016 \cite{UAV} shows that China has more than 3,000 professional small light UAVs for remote sensing applications. This type of UAV exhibits the following five main characteristics.

1) Long flight time. As new energy technology, energy management technology, and lightweight composite material research technology have developed, UAV flight time has been continuously extended.

2) Low comprehensive cost and high technical content. On the one hand, the use of low-cost and lightweight materials reduces the production cost of UAV and remote sensors. On the other hand, the increase of mass users promotes the mass production of components and structural parts, further reducing the production cost of UAV and remote sensors.

3) Small, light-weight, diversified remote sensing cameras. All remote sensing loads on small light UAVs are developed to below 30 kilograms, and optical and infrared loads are even reduced to less than a half kilogram. In addition, multi-angle photography, tilt photography, sensor integration, hyperspectral imaging interference \cite{hong2019cospace}, and other technologies have been used in UAV remote sensing. Commercial high-end cameras have been widely used for professional aerial missions, and popular cameras are used for mass entertainment and general applications.

4) Real-time data transmission. Advances in wireless communication and information compression technology have powerfully impelled image resolution with a higher data rate and longer transmission distance. Almost no-delay data link transmission makes real-time observation possible.

\subsection{Challenges}

Object detection and tracking tasks in the UAV remote sensing video face many challenges, such as image degradation, uneven object intensity, small object size, and real-time problems like perspective specificity, background complexity, scale, and direction diversity problems in satellite and manned aircraft objects.

\begin{itemize}
\item \textbf{\textit{Image degradation problem.}} The load that a mini-UAV platform carries is strictly limited in terms of weight, volume, and power. Rapid movement changes in the external environment (such as light, cloud, fog, rain, etc.) cause aerial images to be fuzzy and noisy, which inevitably leads to image degradation \cite{hong2019augmented}. In addition, high-speed flight or camera rotation also increases the complexity of object detection. Thus, it is necessary to carry out image pre-processing, such as noise reduction, camera distortion correction, etc., to ensure the effectiveness of the object detection model.
\item  \textbf{\textit{Uneven object intensity problem. }}The image acquisition equipment of a UAV typically uses a large aperture, fixed focal, and wide-angle lens. In addition, flexible camera movement results in an uneven density of captured objects. Some of them are densely arranged and overlap many times, so that it is easy to repeat detection. Some are sparse and unevenly distributed, so that it is prone to missed detection. In addition, most objects occupy a small number of pixels, which makes it difficult to separate them from their surroundings.
\item \textbf{\textit{Object size problem.}} UAV remote sensing images can be acquired at different altitudes, yielding photographs containing any size of ground objects. This challenges the classical DL-based method. In addition, ground objects in UAV remote sensing are primarily shown as images with an area smaller than $32\times 32$ pixels. MS COCO dataset\cite{lin2014microsoft} defines small objects due to their less distinct features, yielding more false and missed detection targets.
\item \textbf{\textit{Real-time problem.}} Object detection or tracking in a video obtained by a drone needs to quickly and accurately locate moving ground objects, so real-time processing performance is highly essential.
\end{itemize}

\subsection{Contribution}
Up to now, reviews concerning object detection and tracking from airborne and spaceborne datasets can be found \cite{han2018advanced,li2020object}. For UAV data, several representative surveys have been published in the literature, which include surveys on the UAV image processing and application \cite{2018Mini,rs11121443}, the UAV system\cite{yin2019review}. However, less attention has been paid to the advance of object and tracking techniques both in image and video acquired by UAV. Although reviews in \cite{osco2021review,mittal2020deep,2021A,Ayalew2019ARO,jimaging6080078,electronics10070820} present some DL-based static object detection for UAV image and the one in \cite{hao2018review} presents traditional object tracking for UAV video, there still lacks a complete survey for object and tracking and the most recent advances.

\begin{figure}[!t]
    \centering
    \subfigure{
             \includegraphics[width=0.5\textwidth]{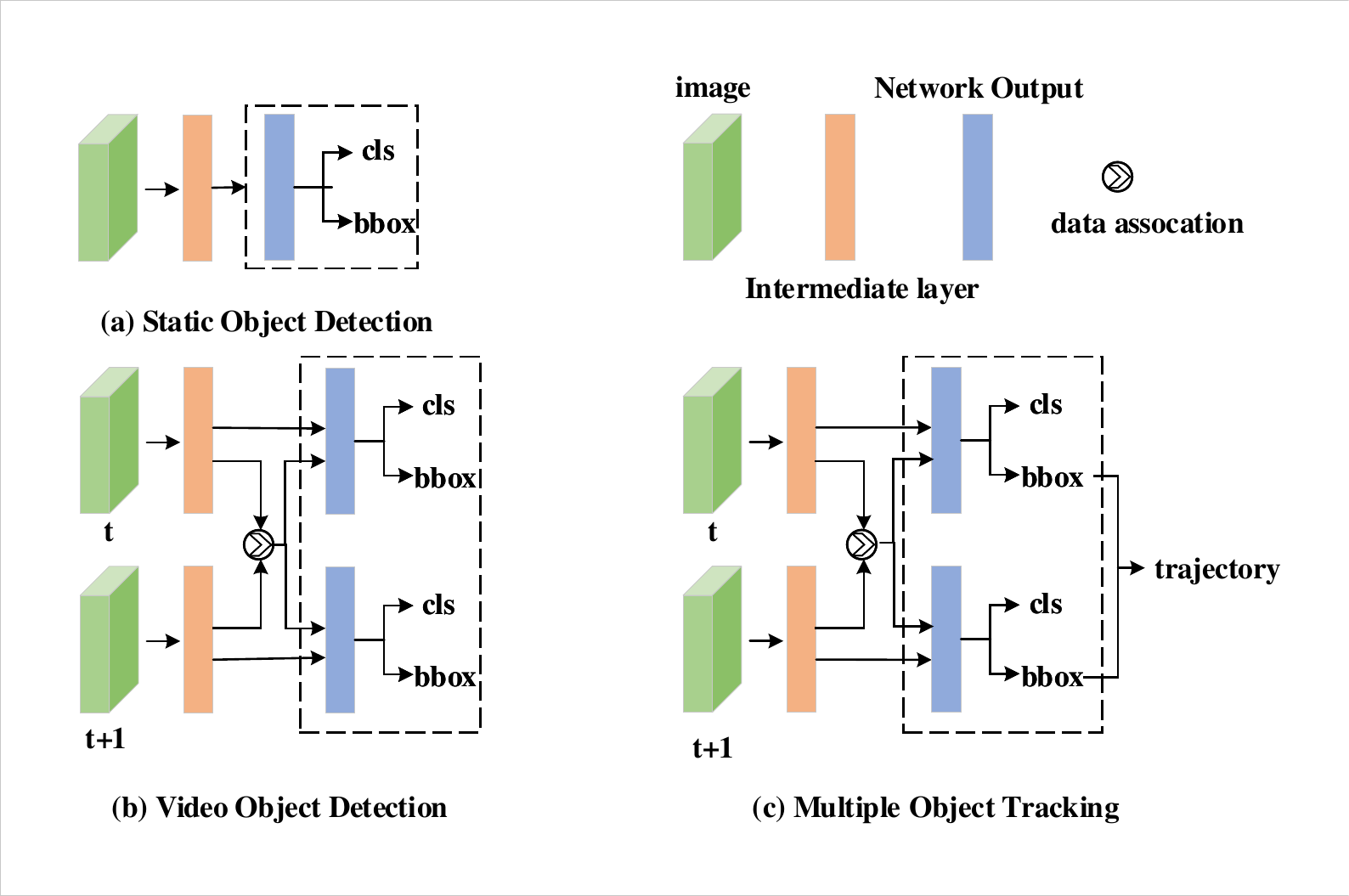}
    }
    \caption{An illustration of three UAV topics based on deep learning methods.}
\label{fig:Com_three_topic}
\end{figure}

Therefore, it is imperative to provide a comprehensive survey of DL-based object detection and tracking for UAV data, focusing on static object detection (SOD), video object detection (VID), and multiple object tracking (MOT). In the following discussion, we limit this review to DL-based methods based on corresponding publications. We hope that this survey will provide readers and practitioners with instructive information. Fig. \ref{fig:Com_three_topic} shows the typical DL-based learning mode for these three research topics. For the choice of the DL method, SOD object detection focuses on detection head design to assign positive and negative samples, such as RPN+ ROI Pooling in Faster RCNN, detection outputs is classification and bounding box. VID and MOT are about UAV video data and the difference between them is how to use temporal information. The former focuses on modifying the missed detection results of the current frame by using temporal context in adjacent frames, while the latter focuses on predicting the trajectory in the next frame to obtain the moving state of objects.

\begin{figure*}[!t]
    \centering
    \includegraphics[width=1.0\textwidth]{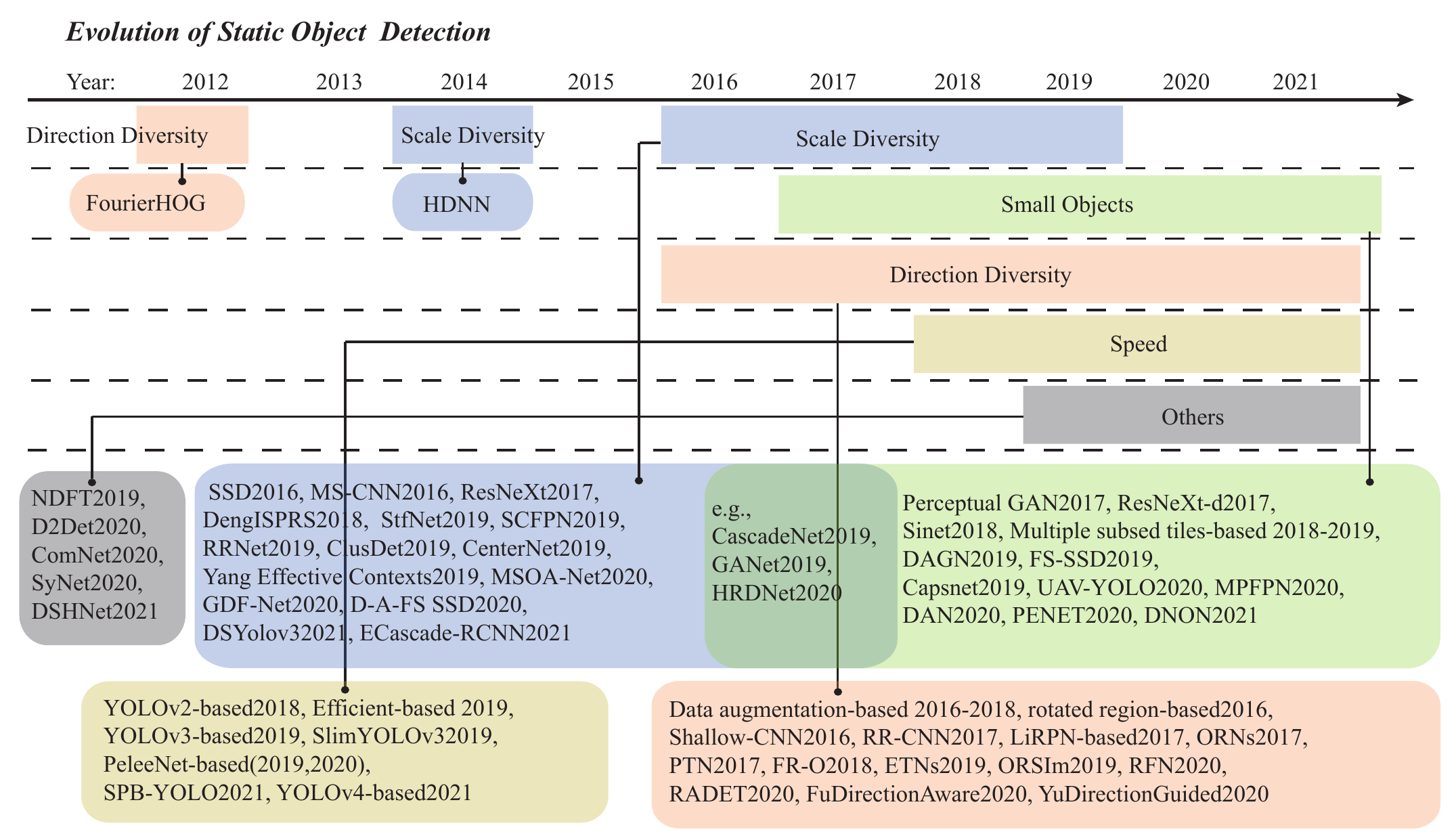}
    \caption{The development of typical methods for UAV static object detection.}
\label{fig: IMA}
\end{figure*}
\begin{figure*}[!t]
    \centering
    \includegraphics[width=1.0\textwidth]{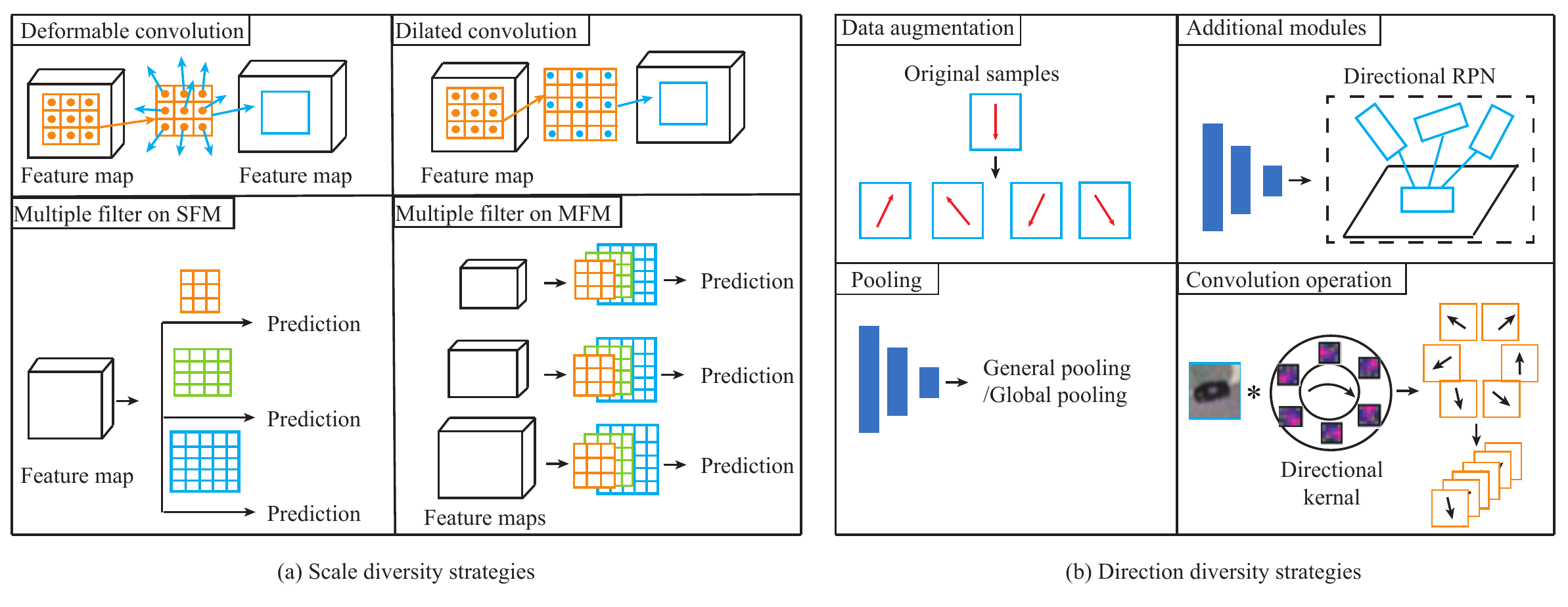}
    \caption{Deep learning-based scale diversity and direction diversity strategies.}
\label{fig: ScaleDirection}
\end{figure*}
\section{Object Detection from UAV-borne Images} \label{sec:VII}

Although deep learning-based object detection methods for UAV remote sensing images are mainly borrowed from traditional digital images in the computer vision community, the limitation of small UAV platform and imaging acquisition condition inevitably causes problems of particularity perspective, complex background, scale and direction diversity, and issues related to small sizes. In the following, some solutions based on DL methods have been summarized according to recent publications. Fig. \ref{fig:VID} shows the development of typical methods for SOD. Among them, some methods specially designed for UAV data are listed in Table.  \ref{tab:SOD}. Other methods that can solve the above problem, but not specifically for UAV data, are briefly introduced in the text. 
The remainder of this section introduces DL-based SOD methods to solve five representative problems, including data processing, scale diversity, small objects, direction diversity and detection speed.

\subsection{Data Processing}
Two types of data processing are typically applied preprocessing before data acquisition and postprocessing after data acquisition. 

The latter is more commonly used in DL-based techniques. Most of the existing UAV-based remote sensing works present an experimental dataset and appropriate data processing techniques \cite{pajares2015overview,adao2017hyperspectral,8833958}, and all of them carry out image postprocessing procedure after image acquisition, such as increasing the number of training samples, enlarging the diversity of sample size and direction, and expanding the illumination change of samples. However, their effectiveness is variable.

Due to the limitation of UAV flight altitude and load, there is inevitably ground object overlapping, coverage, and displacement. Xia et al. \cite{xiang2018mini} took optical cameras as an example, focusing on various difficulties and problems in the process of UAV remote sensing data acquisition, and systematically discussed the key techniques of data processing.

\subsection{Object Detection on Scale Diversity} UAV remote sensing images can be acquired at different altitudes and ground objects can be any size, even for intraclass. Therefore, solutions to scale diversity are cross-referenced in this review. There are two main approaches to solving this problem through deep learning, as illustrated in  Fig. \ref{fig: ScaleDirection}(a). The most commonly used is the multi-scale feature map \cite{wu2018msri}, which is the output of multiple filters on multiple feature maps (MFM) or multiple filters on a single feature map (SFM) \cite{chen2014vehicle,liu2019stfnet,liu2016ssd,cai2016unified,deng2018multi,chen2019rrnet,ECascade-RCNN,DSYolov32021,MSOA-Net2020,ClusDet2019,pailla2019object,yang2019effective}. The other is a dilated/deformable convolution kernel \cite{chen2019object,GDF-Net2020,zhang2019dense,ICIT2020}. It points out that systematic expansion supports the exponential expansion of the receptive field without loss of resolution or coverage. Chen et al. \cite{chen2019object} introduced an extended convolution filter to obtain the ResNeXt-d combination structure on the basis of ResNeXt\cite{xie2017aggregated} architecture, which can expand the receptive field.

\begin{table*}[!t]
\centering
\caption{DL-based Static Object Detection Approaches for UAV exclusive}
\resizebox{1\textwidth}{!}{
\begin{tabular}{l||l|l|l|l|l}
\toprule[1.5pt]
\multicolumn{6}{c}{Static Object Detection}\\ \hline
Reference & Challenge & Dataset Used & Journal/Conf. &Year &Code.link \\
\hline \hline
RRNet\cite{chen2019rrnet}& Small objects, scale variation & VisDrone & 	ICCV Workshops & 2019& \url{https://github.com/ouc-ocean-group/RRNet}\\ 
SlimYOLOv3\cite{zhang2019slimyolov3} &  Real-time & Visdrone & ICCV &2019 & \url{https://github.com/PengyiZhang/SlimYOLOv3.}\\ 
Zhang et al\cite{zhang2019dense} & Small objects & VisDrone & ICCV Workshops &2019&-\\ 
FS-SSD\cite{liang2019small}& Small objects & Stanford Drone  & IEEE TCSVT &2019&-\\
SAMFR\cite{wang2019spatial}&  Scale variation & Visdrone &  ICCV Workshop &2019&-\\ 
ClusDet\cite{ClusDet2019}& Scale variation & VisDrone, UAVDT & ICCV & 2019& \url{https://github.com/fyangneil}\\
CenterNet\cite{pailla2019object}& Scale variation & VisDrone &-& 2019&- \\ 
Yang et al\cite{yang2019effective}& Scale variation & Stanford Drone & IEEE Access & 2019&- \\ 
Wu et al\cite{wu2019DDCLS}&  Real-time & CARPK & DDCLS &2019&-\\
NDFT\cite{wu2019delving} &  UAV-specific nuisances & VisDrone, UAVDT &ICCV&2019&\url{https://github.com/VITA-Group/UAV-NDFT}\\ 
MSOA-Net\cite{MSOA-Net2020}& Scale variation & UVSD & 	Remote Sens. & 2020&- \\ 
GDF-Net\cite{GDF-Net2020}& Scale variation & VisDrone, UAVDT & Remote Sens. & 2020&- \\ 
HRDNet\cite{liu2020HRDNet}& Scale variation & VisDrone & 	CVPR & 2020&- \\
D-A-FS SSD\cite{ICIT2020}\cite{ICIT2020}& Scale variation  & VisDrone & 	ICIT & 2020&- \\ 
UAV-YOLO\cite{liu2020uav} &  Small scale & UAV123, Own & Sensors &2020&- \\  
SyNet\cite{albaba2020synet}&  Class imbalance& VisDrone & ICPR &2020& \url{https://github.com/mertalbaba/SyNet}\\ 
ComNet\cite{li2020comnet}& Blurred edges, low contrast  & Own & IEEE TGRS &2020 &-\\
MPFPN\cite{liu2020small}& Small objects
& VisDrone  & 	IEEE Access &2020&-\\ 
D2Det\cite{cao2020d2det}, & Localization, classification
& UAVDT  & CVPR & 2020& \url{https://github.com/JialeCao001/D2Det.}\\
DAGN\cite{zhang2019dagn}&  Small objects& VEDAI  & 	IEEE GRSL &2020 &-\\ 
GANet\cite{yuanqiang2020guided}&  Small objects & UAVDT, CARPK, PUCPR+  & MM &2020&\url{https://isrc.iscas.ac.cn/gitlab/research/ganet}\\
DAN\cite{jadhav2020aerial} &Dense distribution, small object& Visdrone-det & 	NCC &2020 &-\\ 
Zhang et al\cite{zhang2020coarse} &  Real-time & Stanford drone & Neurocomputing & 2020&-\\
DNOD Eifficientdet\cite{TIAN2021292} & Dense objects,small objects & Visdrone-DET, UAVDT & Neurocomputing &2021 &-\\ 
ECascade-RCNN\cite{ECascade-RCNN}&Scale variation & VisDrone &ICARA &2021&- \\ 
Cas\_RCNN+FPN\cite{youssef2021automatic}& Cost & Visdrone & Transp. Res. Rec. &2021 &- \\
DSYolov3\cite{DSYolov32021}& Scale variation & VisDrone, UAVDT & J. Vis. Commun. Image Represent.& 2021&- \\ 
DSHNet\cite{yu2021towards}& Long-tail distribution & VisDrone, UAVDT & WACV & 2021&\url{https://github.com/we1pingyu/DSHNet}\\  
\bottomrule[1.5pt]
\end{tabular}}
\label{tab:SOD}
\end{table*}

\subsection{Object Detection on Small Objects} 
The UAV flying altitude inevitably causes most objects to be shown in scale diversity, small object size and dense arrangement, resulting in less feature information that can be extracted. Many work deal with the small object detection problem through the same network designing for scale diversity, including RRNet \cite{chen2019rrnet}, HRDNet\cite{liu2020HRDNet}, Cascade network \cite{zhang2019dense}, UAV-YOLO \cite{liu2020uav}, MPFPN \cite{liu2020small}, depthwise-separable attention-guided network (DAGN) \cite{zhang2019dagn}, GANet \cite{yuanqiang2020guided}, and FS-SSD \cite{liang2019small}, ResNeXt-d \cite{chen2019object}, et al. In these methods, accurate feature information learned by small objects is highly important. In addition, some new networks are based on YOLOv4 or Eifficientdet-D7 networks, e.g., DNOD \cite{TIAN2021292}, which are developed to improve the detection speed.

To further improve the distinguish ability of small objects, Li et al. \cite{li2017perceptual} proposed a perceptual GAN to generate a super-resolved representation of small objects. This method uses the structural correlativity of large and small objects to enhance the representation of small objects and give them a similar expression to large objects. Hu et al. \cite{hu2018sinet} found that the structure of small objects after pooling was typically distorted, and proposed a new context-aware region of interest (ROI) pooling method. Chen et al. \cite{chen2019object} proposed an ResNeXt-d combination structure to enhance the perception of small size objects. There are other methods, including changing the anchor information, or cropping multiple subset tiles from the original high-resolution images, to improve the detection performance of small and dense object. Jadhav et al. \cite{jadhav2020aerial} modified the anchor scale and Tang et al. \cite{tang2020penet} designed a coarse anchor-free detector (CPEN) to address dense small object detection. In \cite {ruuvzivcka2018fast, plastiras2018efficient, Unel_2019_CVPR_Workshops}, the authors proposed effective solutions to small object detection from high-resolution images by cropping multiple subset tiles from the original high-resolution images, and learning them through use of a CNN network without degrading the resolution.  


Alternately, the flexible movement of the camera results in an uneven density of captured objects. Tightly packed objects in an image, especially smaller size ground objects, inevitably overlap. Mekhalfi et al. \cite{mekhalfi2019capsule} introduced Capsnets to model the relationship between objects. 

\subsection{Object Detection on Direction Diversity} Objection direction from an optical remote sensing image is related to its actual parking location. The classic CNNs, which benefit from using a rectangular convolution kernel, are sensitive to object direction. Fig. \ref{fig: ScaleDirection}(b) shows four commonly used solutions based on deep learning.

The simplest and most common solution is data augmentation, which can 
make CNNs rotation-invariant though rotation transformation of different angles to extend the training set \cite{cheng2016learning,laptev2016ti,deng2018multi}. Cheng et al. \cite{cheng2016learning} added regularization constraints on the basis of existing CNN architecture to build a rotation-invariant CNN (RICNN). With further researches, Fisher discriminative CNN related rotation-invariant network, called RIFD-CNN, have been proposed to further boost object
detection performance \cite{cheng2016rifd, cheng2018learning}. Laptev et al. \cite{laptev2016ti} added a rotation-invariant pool operator to the penultimate layer of output. The shortcoming of data augmentation is the increased cost of network training and the risk of over fitting. 

Some work directly used additional network modules such as oriented proposal boxes to achieve object detection \cite{liu2016ship,liu2017rotated}, or upgraded the general convolutional filter to a directional channel filter to achieve rotation invariant of texture \cite{marcos2016learning}. The region proposal network (RPN) \cite{li2017rotation,fu2020rotation,yu2020orientation} added to the anchor boxed with multiple angles in order to cover the oriented object. Additionally, inspired by text detection methods \cite{liao2018rotation,liao2018textboxes,jiang2018r,ma2018arbitrary}, Xia et al. \cite{xia2018dota} designed a direction insensitive FR-O network by adding a direction box detection sub-network to Faster RCNN. Li et al. \cite{ li2020radet} proposed RADet to acquire a rotation bounding box with a shape mask. However, the drawback of additional network modules is that the transform parameter estimation is non-adaptive.

Approaches like, Oriented Response Networks (ORNs) \cite{zhou2017oriented}, Polar Transformer Network (PTN) \cite{esteves2017polar}, and Equivariant Transformer Networks (ETNs) \cite{tai2019equivariant}, which were proposed for object detection from natural scenes, also provided a qualitative or qualitative analysis of rotation invariant features. On the basis of these techniques, Zhou et al. \cite{8968762} developed a rotated feature network (RFN) using encoder–encoder architecture for object detection in remote-sensing images. It is worth mentioning that some rotation-invariant methods based on theoretical analysis can cover the intrinsic properties of rotations \cite{schmidt2012learning,wu2019orsim} to extract real rotation- invariant features. Up to now, these methods have have not been widely used in deep learning. 

\subsection{Object Detection on Detection Speed} Limited by flight stability and the load capacity of a micro-mini UAV, the altitude of an airborne remote sensing sensor needs to be adjusted quickly and accurately in real time so that ground objects are always in the monitoring field of vision. Meanwhile, rapid processing and analysis of high-quality remote sensing images obtained by a UAV system in real time is the key for miniature UAV remote sensing.

When considering all the deep learning methods, the most direct way is to choose the right platform, including an ARM, mobile, and embedded platform, or to trim up the classic network architecture to minimize unnecessary channels in the convolutional layer. In \cite {ruuvzivcka2018fast, plastiras2018efficient,wu2019DDCLS}, the authors adopted a YOLO and even a tiny-YOLO network to achieve real-time object detection. Zhang et al. \cite {zhang2019slimyolov3} trimmed the update YOLOv3, and proposed slimYOLOv3, which balanced the number of parameters, memory usage, and inference time to achieve real-time object detection. \cite{Unel_2019_CVPR_Workshops,zhang2020coarse} modified the feature resolution of the lightweight Pelee network \cite{wang2018pelee} to meet real-time needs. Due to the efficiency and power of YOLOv4, many object detection models \cite{9390673, electronics10070820} are based on this network. Ammar et al. \cite{electronics10070820} used YOLOv3 and newly released YOLOv4 to detect vehicles with inference processing speed from 12 fps for $608\times 608$ up to 23 fps for $320\times 320$. Furthermore, Wang et al. \cite{9415214} designed a Strip Bottleneck with YOLO network (SPB-YOLO) based on YOLOv5 for engineering application.

In addition, real-time object detection from images is also a necessary condition for detection-based object detection and object tracking from videos, which will be discussed in Section \ref{sec:VID} and \ref{sec:MOT}, respectively.

\subsection{Object Detection on Others}
Besides the aforementioned main challenges, other problems in object detection in UAV images are addressed, such as Nuisance Disentangled Feature Transform (NDFT) \cite{wu2019delving} for a large number of fine-grained domains, D2Det \cite{cao2020d2det} for precise localization and accurate classification, combinational neural network (ComNet) \cite{li2020comnet} for blurred edges and low contrast, an ensemble network (SyNet) \cite{albaba2020synet} for class imbalance problem and the scaling problem, and Dual
Sampler and Head detection Network (DSHNet) for long-tail distribution.

\begin{figure*}[!t]
    \centering\includegraphics[width=1.0\textwidth]{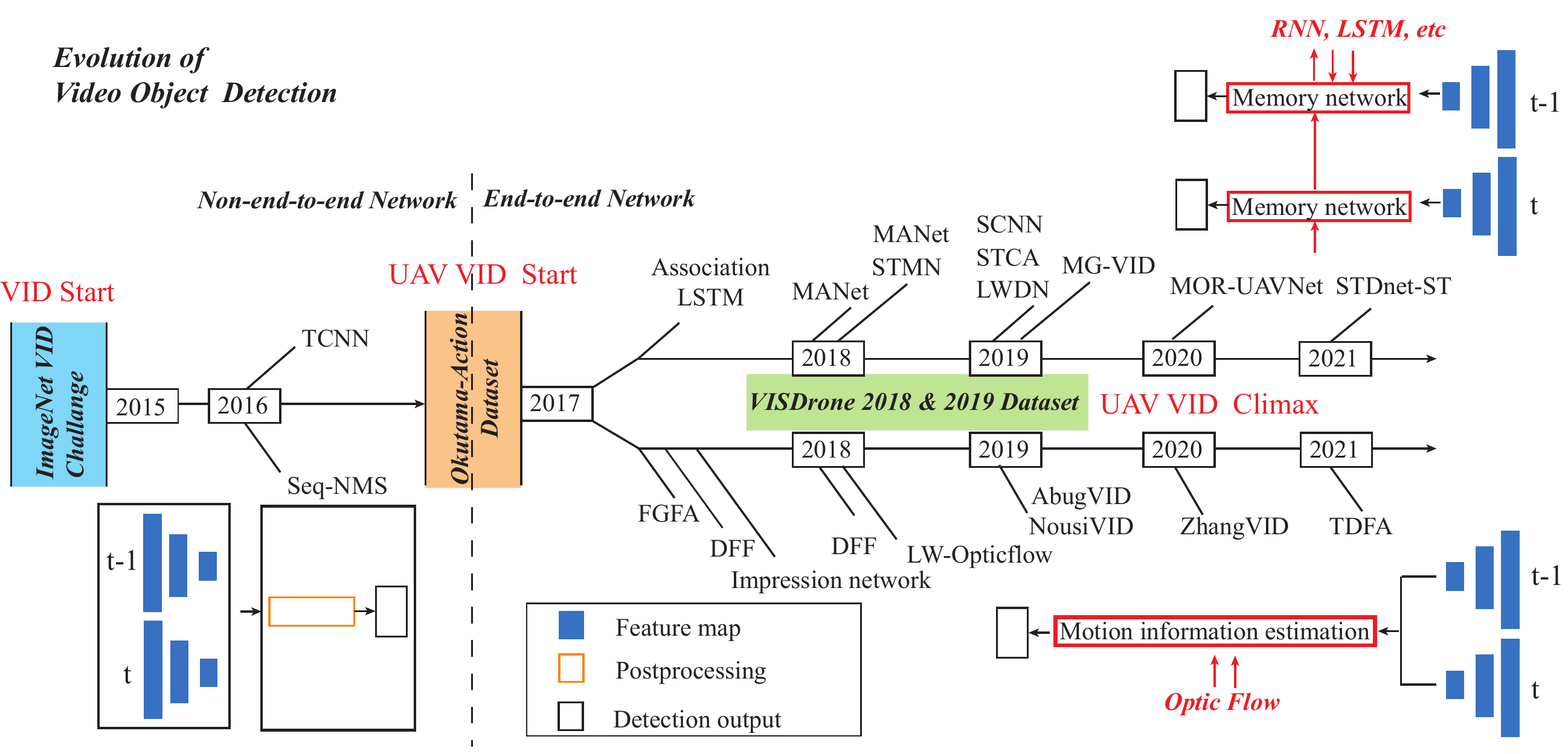}
    \caption{The development of typical methods for UAV object detection from video.}
\label{fig:VID}
\end{figure*}

\begin{table*}[!t]
\centering
\caption{DL-based Video Object Detection Approaches for UAV exclusive}
\resizebox{1\textwidth}{!}{
\begin{tabular}{l||l|l|l|l|l}
\toprule[1.5pt]
\multicolumn{6}{c}{Video Object Detection}\\ \hline
Reference & Challenge & Dataset Used & Journal/Conf. &Year &Code.link \\
\hline \hline
STCA\cite{Pi_2019_ICCV}&Defocus, motion blur, occlusion &VisDrone-VID  &ICCV Workshop  &2019&-\\ 
SCNN\cite{wang2019scnn} & Temporal and contextual correlation&DAC & AAAI &2019  &-\\ 
Nousi et al.\cite{nousi2019embedded} &Real-time &own-recorded  & RCAR  &2019 &-\\
Abughalieh et al\cite{abughalieh2019video}&   Varying resolutions& Own &Multimed. Tools. Appl.   &2019 &-\\
Zhang et al\cite{zhang2020drone} & Appearance deterioration, occlusion, motion blur&VisDrone-VID & MIPR  &2020&-\\ 
MOR-UAVNet\cite{mandal2020mor}&   Moving object &MOR-UAV &MM&2020&\url{https://visionintelligence.github.io/Datasets.html}\\ 
TDFA\cite{xie2021two} &Small-scale&Okutama, VisDrone-VID& 	Multidim Syst Sign P & 2021 &-\\ 
STDnet-ST\cite{bosquet2021stdnet}&Small object
& USC-GRAD-STDdb,UAVDT,VisDrone-VID  & 	 PR  & 2021&-\\
\bottomrule[1.5pt]
\end{tabular}}
\label{tab:VID}
\end{table*}

\section{Object Detection from UAV-borne Video}\label{sec:VID}
Video object detection (VID) becomes a hot topic after ImageNet VID challenge 2015. It is widely used on UAV data until 2017, and also brings some new challenges, e.g., camera change and motion blur in a drone platforms. In the following, some solutions based on DL methods are summarized according to recent publications. Fig. \ref{fig:VID} shows the development of typical methods for VID. Among them, methods specially designed for UAV data are listed in Table \ref{tab:VID}. Other methods that can solve the above problem, but not specifically for UAV data, are described in the text.


The main steps of VID are summarized below.

\begin{itemize}
\item Single frame image object detection: Static object detection or object detection from images. Each frame in the video is an independent image, and the object detection from the image can be achieved by using a method in Section \ref{sec:VII}.
\item Detection results amendment: The above missed detection results are compensated by temporal information and context information of the video.
\end{itemize}


The early mainstream method for VID is a multi-stage pipeline method, such as tubeless with convolutional neural network (TCNN) \cite{kang2016object, kang2017t} and sequence non-maximum suppression (Seq-NMS) \cite{han2016seq}, which is object detection from each frame where the modified detection results using a temporal context are performed separately. 
With the depth of research, many approaches have started to cast VID as a classic object detection problem. For example,  
a network model of feature enhancement module integration called SSD with comprehensive feature enhancement (CFE-SSDv2) \cite{zhao2018comprehensive} has been proposed to improve the accuracy of small size VID. F-SSD \cite{Pi_2019_ICCV}, based on SSD and FCOS, improved the robustness of the model through decision fusion of each frame detection result. EODST\cite{danelljan2017eco}, based on SSD, adopted ECO tracking methods to associate object detection from a single frame. Similarly, benefiting from some advanced detectors, like HRDet \cite{sun2019deep}, Cascade R-CNN \cite{cai2018cascade}, CenterNet \cite{zhou2019objects}, RetinaNet \cite{lin2017focal}, and FPN \cite{lin2017feature}, statistical convolutional neural network (SCNN) \cite{wang2019scnn},  several networks specifically developed for VID have been proposed. Some literature focus on the performance and real-time of these networks, so as to develop them on mobile \cite{abughalieh2019video} or embedded systems \cite{nousi2019embedded}. However, these methods are difficult to cover the context information for video. Although there are some methods to integrate Spatio-temporal information, e.g., Spatio-temporal neural network built on STDnet (STDnet-ST) \cite{bosquet2021stdnet}, the problems of missed and false inspection still persist.

The remainder of this section introduces three mainstream DL-based VID methods, including optical flow-based network, memory network-based network and tracking-based network, which integrate temporal context information into the DL-based methods to yield a better detection performance of VID and correct false alarms and missed detection. 


\subsection{Optical Flow-based Network} 
In order to build the relationship between consecutive frames, some researchers estimate motion information. The most commonly used motion estimation method is optical flow. 

\cite{zhang2020drone} and \cite{xie2021two} used the effective CNN model for optical flow (PWC-Net) \cite{sun2018pwc} method and spatial pyramid network (SPyNet) \cite{8099774} to obtain the motion information of two neighbor frames, respectively. Zhu et al. \cite{zhu2017deep} designed fusion feature maps to achieve VID using deep feature flow (DFF) by learning the feature maps of key frames using feature extracting and of non-key frames using FlowNet. FlowNet was 11.8 times faster than Mobilenet, and even the smallest FlowNet-Xception was 1.6 times faster. The flow guided feature aggregation (FGFA) \cite {zhu2017flow} proposed by the MSRA visual computing group is also an early attempt based on optical flow. FGFA enhances the features of each frame by aggregating the features of multiple frames, finally using FlowNet to warp the features to solve video degradation. While FGFA is helpful for medium and fast speed VID, it is less effective for slow-speed VID. Subsequently, FGFA+ achieved better results by merging several data expansion strategies. Ref. \cite {hetang2017impression} proposed an impression network, that can perform multi-frame feature fusion between sparse key frames, solving problems like defocus, motion, blur, and other issues in VID, while balancing detection speed and accuracy. Built upon \cite{zhu2017flow,zhu2017deep}, Zhu et al. \cite{zhu2018towards} adapted the flow network to learn multi-frame features and estimate cross-frame motion. Zhu et al. \cite{zhu2018towards2} subsequently designed a more lightweight optical flow network on mobiles. The entire network is trained end-to-end, reaching a mean average precision (MAP) of 60.2 in VID, and running to a speed of 25 frames on a Huawei Mate 8 cellphone. Due to a large amount of optical flow calculation using multiple frames, the network cannot perform back propagation revision during the training phase.

\subsection{Memory Networks-based Network} 
Since a video sequence has a strong long-term correlation, researchers introduced a memory network to fully learn the time information in a video sequence, such as a recurrent neural network (RNN) \cite{hang2019cascaded}, long short-term memory (LSTM), and gated recurrent unit (GRU). 

In \cite{lu2017online}, Lu et al. proposed an association LSTM that fundamentally modeled object association between consecutive frames, and prompted LSTM to supply high quality association features. Refs. \cite {hetang2017impression} and \cite{liu2019looking} both used ConvLSTM for efficient fusion of multi-frame features, improving video object detection accuracy while ensuring timeliness. In particular, \cite {liu2019looking} developed a new cross framework that used two feature extractors to run on different frames to improve the robustness of detectors. Liu et al \cite{liu2018mobile} proposed an inter woven recurrent-convolutional architecture by designing the Bottleneck-LSTM layer to ensure real-time detection. Inspired by \cite{zhu2018towards} and \cite {hetang2017impression}, Jiang et al. \cite{jiang2019video} adopted a brain-inspired memory mechanism to design a locally weighted deformable neighbors method for video object detection. Tripathi et al. \cite{tripathi2016context} trained RNN through the content information of adjacent frames to optimize VID. Unlike the motion information learning of adjacent frames, Xiao \textit{et al} \cite{xiao2018video} proposed a spatio-temporal memory network (STMN) to model and align the long-term sequence appearance and motion dynamics of objects in an end-to-end manner by learning multiple frames information. Wang et al. \cite{xiao2018video} proposed a motion-aware network (MANet) to directly learn motion information over a long period of time by fusing multiple frame features. 

\subsection{Tracking-based Network} In view of the high similarity between VID and object tracking in video discussed in the next section, there are still some methods to achieve VID by means of a tracking method \cite{Pi_2019_ICCV} or to achieve object detection and tracking at the same time \cite{luo2019detect}. \cite{Pi_2019_ICCV} proposed a novel spatial and temporal context-aware approach based on tracking for drone-based video object detection. In \cite{luo2019detect}, the authors designed a scheduler network as a generalization of siamese trackers determined to detect or track at a certain frame. Actually, detection and tracking always coexist in actual scenarios.

\begin{figure}[!t]
    \centering\includegraphics[width=0.48\textwidth]{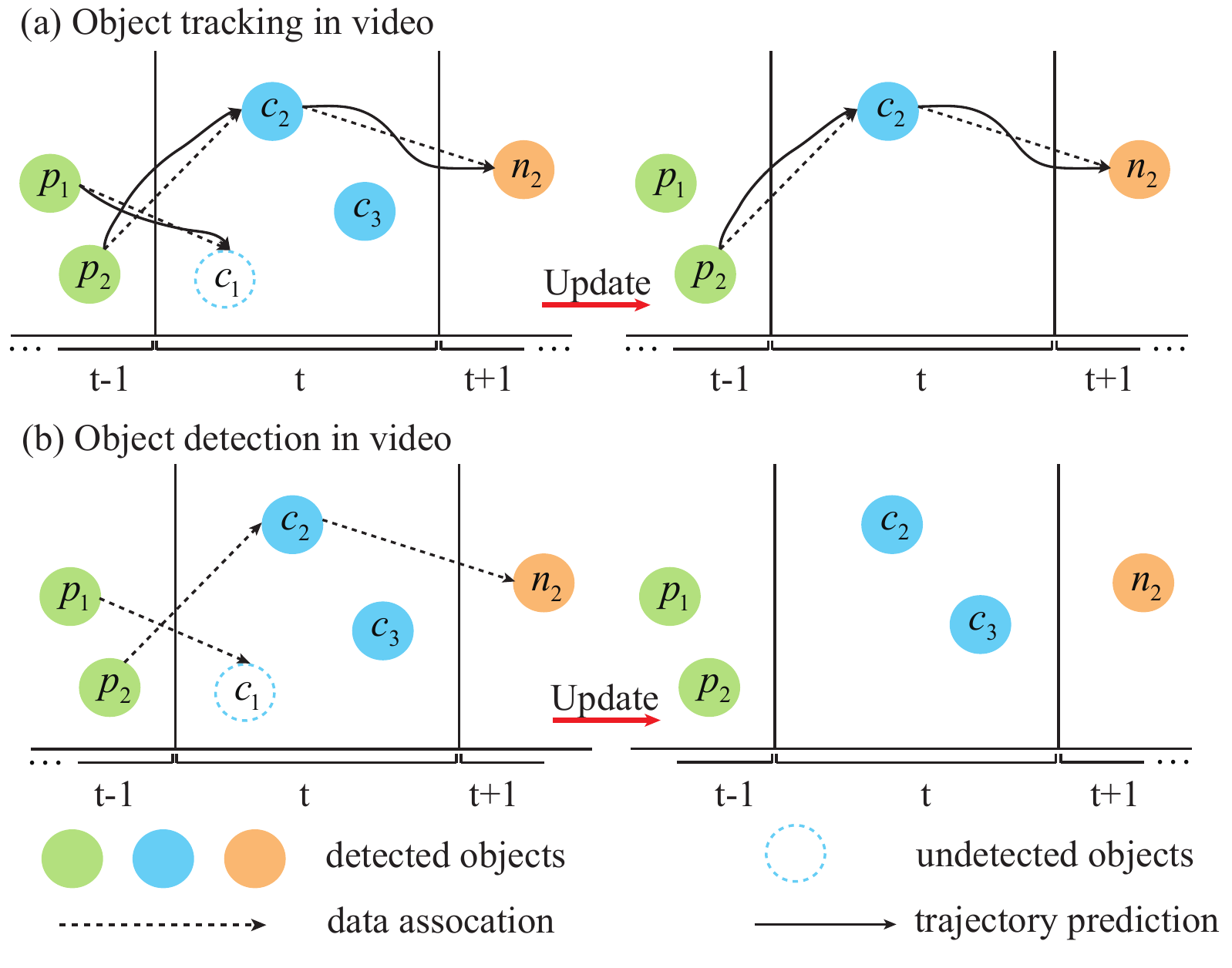}
    \caption{An illustration of the difference between VID and MOT in up/down frames. }
\label{fig:VID-MOT}
\end{figure}
\begin{figure*}[!t]
    \centering\includegraphics[width=1.0\textwidth]{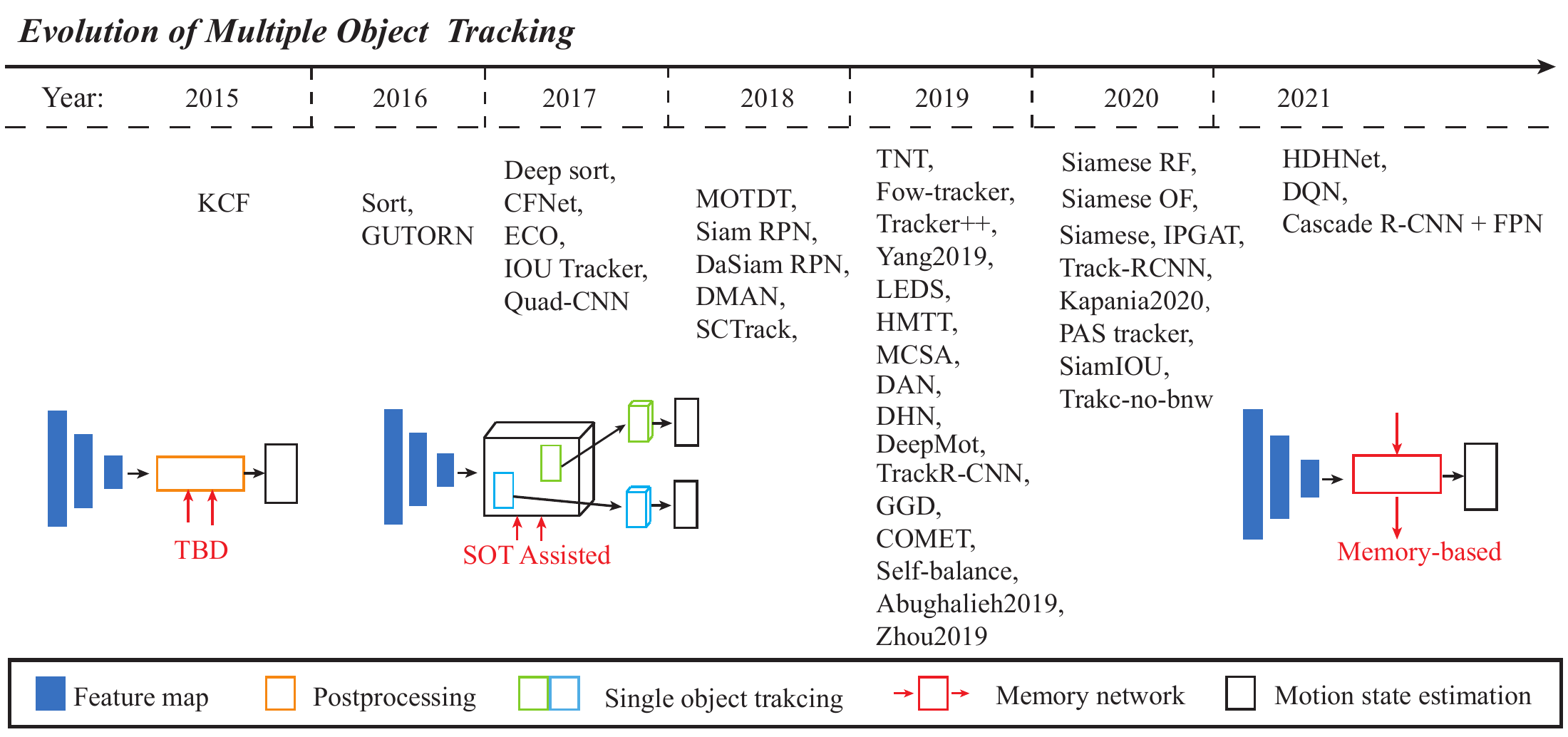}
    \caption{The development of typical methods for UAV object tracking from video.}
\label{fig:MOT}
\end{figure*}

\begin{table*}[!t]
\centering
\caption{DL-based Multiple Object Tracking Approaches for UAV exclusive}
\resizebox{1\textwidth}{!}{
\begin{tabular}{l||l|l|l|l|l}
\toprule[1.5pt]
\multicolumn{6}{c}{Multiple Object Tracking} \\
\hline 
Reference & Challenge & Dataset Used & Journal/Proc. & Year & Code/Link\\
\hline \hline
Deep SORT \cite{wojke2017simple} & Occlusion&VisDrone-MOT &	
ICIP  & 2017&\url{https://github.com/nwojke/deep_sort}\\ 
SCTrack \cite{al2018multi} &  Missed detection, occlusions&VisDrone & AVSS   &2018&-\\ 
Zhou et al\cite{zhou2019uav} & Occlusion & VisDrone-MOT & Comput. Electr. Eng. & 2019 &- \\
OSIM\cite{rs11182155}  & Orientation, scale & UAVDT &Remote Sens. & 2019 & -\\
Flow-tracker\cite{li2019multiple} &  ID Switches, error detection &VisDrone-MOT &ICCV  & 2019&-\\ 
TNT\cite{zhang2019eye}  &   Camera motion, occlusion, pose variation &VisDrone-MOT, Own &ACM-MM  &2019&- \\
HMTT\cite{pan2019multi}  &  Target motion, shape, appearance changes &VisDrone-MOT & ICCV  &2019 &-\\ 
Yang et al\cite{yang2019multiple}  &  Target position changes  &Own& RS  &2019& \url{https://frank804.github.io/} \\  
GGD\cite{ardo2019multi} &  False alarms, missed detections&VisDrone-MOT &ICCV    & 2019&\url{https://github.com/hakanardo/ggdtrack}\\
COMET\cite{marvasti2020comet} &Small object &UAVDT, VisDrone-MOT, Small-90& 		
ICCV   &2019 &-\\
Self-balance\cite{yu2019self} & Appearance, motion &UAVDT & Multimedia Asia  &2019&- \\  
Abughalieh et al\cite{abughalieh2019video}&  Low detailed targets & DARPA, VIVID, Own& 	Multimed. Tools.Appl.   & 2019&-\\ 
Tracktor++\cite{bergmann2019tracking}  & Occlusions, crowded scenes &VisDrone-MOT&ICCV  &2019&\url{ https://git.io/fjQr8}\\
IPGAT\cite{yu2020conditional}  &Small object, appearance unreliable&UAVDT, Stanford Drone & PRL & 2020&- \\
Kapania et al\cite{kapania2020multi}& Real-time &VisDrone-MOT & AIMS  & 2020&-\\ 
PAS tracker\cite{stadler2020pas} & False detections &VisDrone-MOT & ECCV  & 2020&-\\ 
DAN\cite{jadhav2020aerial}  & Dense distribution, small object&VisDrone-MOT& NCC &2020&-\\ 
DQN\cite{dike2021robust}  & Small target  &UAVDT &  Electronics  & 2021&-\\ 
Cas\_RCNN+ FPN\cite{youssef2021automatic}  & Complex background & VisDrone-MOT & Transp. Res. Rec. & 2021 &- \\
HDHNet\cite{huang2021multiple} & Small object,class imbalance &VisDrone-MOT & Multimed. Tools. Appl. &2021& -\\ 
\bottomrule[1.5pt]
\end{tabular}}
\label{tab:MOT}
\end{table*}

\section{Multiple Object Tracking from UAV-borne Video} \label{sec:MOT}

Multiple object tracking (MOT) for UAV video attract increasing research interest in recent years due to the flexibility of the camera in the drone platform. The popular DL-based MOT methods are not usually optimal for drone video data, due to new challenges, e.g., large viewpoint change and scales in drone platforms. Fig. \ref{fig:VID-MOT} shows a brief process to clarify the differences between VID and MOT. Both VID (Section \ref{sec:VID}) and MOT need accurate object location,and the difference in MOT lies in predicting the trajectory in the next frame, in order to obtain the moving state of objects. In contrast, VID only needs to modify the detection results of the current frame by using temporal context in adjacent frames. In the following, DL-based solutions are summarized according to recently published literature. Fig. \ref{fig:MOT} shows the development of typical methods for MOT. Among them, methods specially designed for UAV data are listed in Table \ref{tab:MOT}. Others methods that can solve the above problem, but not specifically for UAV data, are described directly in the paragraph. The remainder of this section introduce three mainstream DL-based MOT methods, i.e., tracking-by-detection, single object tracking assisted method, and memory networks.

\subsection{Tracking-by-Detection} \label{subsec:TBD}

Tracking-by-Detection (TBD) is the mainstream method of MOT \cite{yu2020conditional,bae2014robust,xiang2015learning,wojke2017simple,pirsiavash2011globally,dicle2013way,dicle2013way,bochinski2017high}. The main steps of TBDs are to first detect all objects of interests for the current frame, and then perform data associated with the previous frame for tracking. This method has the virtue of tracking newly arising objects in the whole video, but detection accuracy has a decisive effect on tracking results. In the TBD method, MOT is considered a data-dependent problem.

The commonly used TBD is CMOT \cite{bae2014robust}, MDP \cite{xiang2015learning}, SORT \cite{bewley2016simple} and DSORT \cite{wojke2017simple,kapania2020multi,jadhav2020aerial}, GOG \cite{pirsiavash2011globally}, CEM \cite{milan2013continuous}, SMOT \cite{dicle2013way}, and IOUT \cite{bochinski2017high,li2019multiple}. For these methods, DL is only responsible for object detection, and traditional data-related methods are for data association. Recently, many learning-based data association approaches have been proposed. For example, Schulter et al. \cite{schulter2017deep} designed an end-to-end network to solve the association problem. Son et al. \cite{son2017multi} proposed a quadruplet convolutional neural network (Quad-CNN) with learning data association across frames by quadruplet losses. Feichtenhofer et al. \cite{feichtenhofer2017detect} introduced correlation features and produced data association cross frames by linking the frame-level detection, which could simultaneously achieve object detection and tracking. Sun et al. \cite{sun2019deep} adopted a depth network to realize end-to-end feature extraction and data association. Jadhav et al. \cite{jadhav2019aerial} proposed multiple object tracking methods by training a custom deep association network. Zhang et al. \cite{zhang2019eye} developed a UAV tracking system, which is an integration of RetinaNet and TrackletNet Tracker (TNT). Huang et al. \cite{huang2021multiple} proposed a hierarchical deep high-resolution network (HDHNet) to achieve an end-to-end online MOT system. Stadler et al. \cite{stadler2020pas} proposed a PAS tracker that employs a novel similarity measure and Cascade RCNN to make full use of object representations. Yang et al. \cite{yang2019multiple} designed dense-optical-flow-trajectory voting to measure the similarity of objects in adjacent frames, and integrated YOLOv3 to realize MOT.

Another way to optimize the track association is Siamese network \cite{chopra2005learning}, which is a similarity measurement method that is especially suitable for object classification when there are more object classes but small quantities in each class. It has been widely applied in multiple object tracking \cite{lee2020online,jin2020online,shuai2020multi,bae2017confidence,leal2016learning,wang2016joint}. For example, LEE et al. \cite{lee2020online} proposed an on online object tracking using rule distillated Siamese random forest. Jin et al. \cite{jin2020online} proposed online MOT with Siamese network and optical flow (Siamese-OF). Shuai et al. \cite{shuai2020multi} proposed MOT with Siamese Track-RCNN. Bea et al. \cite{bae2017confidence} proposed an updated Siamese network to learn discriminative deep feature representations for MOT. Leal-Taix\'e et al. \cite{leal2016learning} developed a multi-modal MOT method by learning the local features of RGB images and optical flow maps using a Siamese network. Al-Shakarji et al. \cite{al2018multi} designed a time-efficient detection-based multi-object tracking system using a three step cascaded data association scheme. Dike et al. \cite{dike2021robust} proposed a quadruplet network to track prediction objects from crowded environments. Yu et al. \cite{yu2019self} proposed a self-balance method integrating appearance similarity and motion consistency. Youssef et al. \cite{youssef2021automatic} achieve MOT by cascade region-based convolutional neural networks and feature pyramid networks.

It should be noted that if we directly use video data acquired by UAV during the flight for MOT, the detection result often contain high noise, false alarm's and missed detection due to changes in the UAV aircraft's motion, the inevitable "jitter'', and ambient light. Therefore, it is necessary to pre-process the UAV video. In addition, Tracking-by-Detection would fail to efficiently match when the front and back frames of the object in the video move too fast. 

\subsection{Single Object Tracking Assisted Multiple Object Tracking}
Trajectory prediction can address the failings of Tracking-by-Detection identified above well, and the most commonly used method is the single object tracking (SOT) assisted method \cite{li2020real,yan2012track, xiang2015learning, chu2017online, sadeghian2017tracking, pan2019multi}. With significant progress in this approach recently, SOT has been successfully applied to complex scenes \cite{bertinetto2016staple,fan2017parallel,fan2017sanet},  but directly applying SOT to MOT would encounter calculation inefficiency and tracking drift caused by occlusion. For this reason, Pan et al.\cite{pan2019multi} propose a hierarchical multi-target tracker (HMTT) incorporating SOT and Kalman filtering to improve the MOT performance. Li et al. \cite{li2020real} designed an multiple vehicle tracking approach to effectively integrate SOT based forward position prediction with IOUT to enhance the detection results in the association phase. Yan et al. \cite{yan2012track} associated detector and SOT trackers as candidate objects, and then candidates were selected through an ensemble framework. Xiang et al. \cite{xiang2015learning} adopted the Markov Decision Processes (MDP) method to track objects in a tracked state with optical flow. Chu et al. \cite{chu2017online} treated all detection output as SOT proposals, and designed MOT network architecture by considering multiple objective interactions, yielding a significant improvement for MOT. Ref. \cite{chu2019online} proposed a novel instance-aware tracker to effectively integrate SOT in to MOT. In \cite{feng2019multi}, the authors adopted a Siamese-RPN \cite{li2018high} SOT tracker and re-identification (ReID) network to extract short-term and long-term clues, respectively. Better data association method called Switcher-aware classification (SAC) was then proposed to improve the tracking results while solving the offset problem. In the above methods, the SOT tracker is independent of data association, which raises a potential issue that the two steps do not collaborate well to reinforce each other. To this end, Zhu et al. \cite{zhu2018online} proposed Dual Matching Attention Networks (DMAN) to deal with intra-class distractors and frequent interactions between objects though integrating a unified framework by single object ECO tracking and data association. 

In addition, for real-time analysis of the SOT-assisted method, offline-trained SOT trackers like the Siamese-RPN can achieve high-speed accuracy of more than 80 frames per second, while an online SOT update consumes a lot of CPU resources.

\subsection{Multiple Object Tracking Based on Memory Networks}
Similar to the VID, MOT may judge new object status through historical trajectory information. Therefore, it is a feasible framework for designing a network structure that can memorize historical information and learn matching similarity measurement based on this historical information to enhance the performance of MOT \cite{milan2017online}. Among all the RNNs, the LSTM network has shown reliable performance on many sequence problems, and can overcome the gradient disappearance and explosion problems of standard RNNs. The special structures of LSTM enable it to remember information for a long time. Recently, some methods \cite{li2016learning, alahi2016social,liang2018lstm} employing have achieved impressive performances by LSTM networks. Milan et al. \cite{milan2017online} trained an end-to-end LSTM network for online MOT. Sadeghian et al. \cite{sadeghian2017tracking} integrated appearance, action, and interaction cues into a unified RNN, and designed feature fusion based on LSTM to express motion interaction, so as to learn the matching similarity between the trajectory history information and the current detection. After designing and analyzing each gate function in LSTM, Kim et al. \cite{kim2018multi} proposed a novel RNN model called bilinear LSTM based on multiplication so as to improve the learning ability of long-term appearance models. Yu et al. \cite{yu2020conditional} estimated the individual motion and global motion by LSTM and Siamese network.
In \cite{feng2019multi}, the short-term clues obtained by the Siamese-RPN network and the long-term clues obtained by ReID were introduced to meet complex scenarios and achieved state-of-the-art tracking performance.

\subsection{Multiple Object Tracking Based on Others}
Besides the aforementioned methods, other methods for multiple object tracking are also available, such as generalized graph differences (GGD) for network flow optimization \cite{ardo2019multi} with an efficient representation of differences between graphs, context-aware IoU-guided tracker (COMET) \cite{marvasti2020comet} with offline proposal generation and multitask two-stream network. There is also literature focusing on designing MOT patrol \cite{zhou2019uav} or mobile \cite{abughalieh2019video} systems for UAV video.

\section{UAV-based Benchmark Dataset} \label{sec:Data}
With the development of data-driven deep learning methods, researchers have made a lot of contributions to develop a variety of reference datasets for object detection (including images and videos) and tracking in UAV remote sensing, to help further study and performance comparison. In this section, we have reviewed some of the most commonly used open and classic UAV-based remote sensing datasets for detection and tracking. 
\begin{table*}[!t]
\caption{Comparison of Current State-of-the-Art UAV Benchmarks and Datasets. The tasks SOD, VID, SOT, and MOT stands for object detection from images, object detection from video, single object tracking, and multiple objects tracking respectively. S: Single camera view, M: Multiple camera view, C\_View: Camera View}
\centering
\resizebox{1\textwidth}{!}{
\begin{tabular}{l|c|c|c|c|c|c|c|c|c|c}
\toprule[1.5pt]
\multirow{3}{*}{\textbf{Object Detection from Image}}&\multicolumn{9}{c}{Attributes}\\
\cline{2-11} 
&\multirow{2}{*}{Modality} &\multirow{2}{*}{Images} & \multirow{2}{*}{Boxes}& \multirow{2}{*}{Tasks} & \multirow{2}{*}{Image Size} & \multirow{2}{*}{Annotation } &\multirow{2}{*}{Occlusion} &\multirow{2}{*}{Weather} & \multirow{2}{*}{C\_View} &\multirow{2}{*}{Year}\\
& & &  &  &  & &  &  &\\ \hline
CARPK\cite{Hsieh2017ICCV} &RGB &1.5k &90k &SOD & $1280\times 720$ & HBB & & & S & 2017\\ \hline 
UAVDT\cite{du2018unmanned} &RGB &40k &841.5k & SOD & $1080\times 540$ & HBB &$\surd$ & $\surd$ & M& 2018\\ \hline
DAC-SDC \cite{xu2019dac} &RGB & 150k & - &SOD & $640\times 360$ & HBB &&  & M& 2019\\ \hline
VisDrone-2018\cite{zhu2018vision} &RGB & 40.0k & 183.3k &SOD & $3840\times 2160$ & HBB &$\surd$ & $\surd$ & M& 2018\\ \hline
VisDrone-2019\cite{zhu2020vision} &RGB & 261.9k & 2.6m &SOD & $3840\times 2160$ & HBB &$\surd$ & $\surd$ & M& 2019\\
\hline 
DroneVehicle \cite{zhu2020drone} &RGB + Infrared & 31.064k & 88.3k &SOD & $840\times 712$ & OBB &  & $\surd$ & M& 2020\\ \hline
AU-AIR \cite{bozcan2020air} &Multi-modal & 32.823k & - &SOD & $1920\times 1080$ & HBB &  & $\surd$ & M& 2020\\ \hline
BIRDSAI \cite{bondi2020birdsai} &Thermal-IR & - & 270k &SOD& $640\times 480$ & HBB &  &  & M& 2020\\ \hline
UVSD \cite{MSOA-Net2020} &RGB & 5.8k &  &SOD& $960\times 540$ to $5280\times 2970$ & HBB/OBB &  &  & M& 2020\\ 
\hline
\multirow{2}{*}{MOHR\cite{zhang2021empirical}} &\multirow{2}{*}{RGB} & \multirow{2}{*}{-} & \multirow{2}{*}{90k} &\multirow{2}{*}{SOD}& $5482\times3078$/$7360\times 4912$ & \multirow{2}{*}{HBB} &  &  & \multirow{2}{*}{M}& \multirow{2}{*}{2021}\\ &&&&&$8688\times 5792$&&&&& \\ \hline
\hline
\multirow{3}{*}{\textbf{Object Detection from Video}}&\multicolumn{9}{c}{Attributes}\\
\cline{2-11} 
&\multirow{2}{*}{Modality} &\multirow{2}{*}{Images} & \multirow{2}{*}{Boxes}& \multirow{2}{*}{Tasks} & \multirow{2}{*}{Image Size} & \multirow{2}{*}{Annotation } &\multirow{2}{*}{Occlusion} &\multirow{2}{*}{Weather} & \multirow{2}{*}{C\_View} &\multirow{2}{*}{Year}\\
& & &  &  &  & &  &  &\\ \hline
Okutama-Action\cite{barekatain2017okutama} &RGB &77.4k &422.1k &VID & $3840\times 2160$ & HBB & & &M & 2017\\ \hline
UAVDT\cite{du2018unmanned} &RGB &80k &841.5k & VID & $1080\times 540$ & HBB &$\surd$ & $\surd$ & M& 2018\\ \hline
VisDrone-2018\cite{zhu2018vision} &RGB & 40.0k & 183.3k &VID & $3840\times 2160$ & HBB &$\surd$ & $\surd$ & M& 2018\\ \hline
VisDrone-2019\cite{zhu2020vision} &RGB & 261.9k & 2.6m &VID & $3840\times 2160$ & HBB &$\surd$ & $\surd$ & M& 2019\\ \hline 
MOR-UAV \cite{mandal2020mor} &RGB & 10k & 90k &VID & $1280\times 720$/$1920\times 1080$ & HBB & $\surd$ & $\surd$ &  M & 2020\\ \hline \hline
\multirow{3}{*}{\textbf{Object Tracking from Image}}&\multicolumn{9}{c}{Attributes}\\
\cline{2-11} 
&\multirow{2}{*}{Modality} &\multirow{2}{*}{Images} & \multirow{2}{*}{Boxes}& \multirow{2}{*}{Tasks} & \multirow{2}{*}{Image Size} & \multirow{2}{*}{Annotation } &\multirow{2}{*}{Occlusion} &\multirow{2}{*}{Weather} & \multirow{2}{*}{C\_View} &\multirow{2}{*}{Year}\\
& & &  &  &  & &  &  &\\ \hline
UAV123\cite{mueller2016benchmark} &RGB &110k &110k &SOT & $720\times 720$ & HBB & & & M & 2016\\ \hline
DTB70\cite{li2017visual} &RGB & - & - & SOT & $1280\times 720$ & HBB & & & M & 2017\\ \hline
Stanford\cite{robicquet2016learning} &RGB &929.5k &19.5k &MOT & $1417\times 2019$ & HBB & $\surd$ & & S & 2016\\ \hline
UAVDT\cite{du2018unmanned} &RGB &80k &841.5k & MOT & $1080\times 540$ & HBB &$\surd$ & $\surd$ & M & 2018\\ \hline
VisDrone-2018\cite{zhu2018vision} &RGB & 40.0k & 183.3k & MOT & $3840\times 2160$ & HBB &$\surd$ & $\surd$ & M& 2018\\ \hline
VisDrone-2019\cite{zhu2020vision} &RGB & 261.9k & 2.6m &MOT & $3840\times 2160$ & HBB &$\surd$ & $\surd$ & M& 2019\\ \hline 
BIRDSAI\cite{bondi2020birdsai} &Thermal-IR & 162k & 270k &MOT& $640\times 480$ & HBB &  &  & M& 2020\\ 
\bottomrule[1.5pt]
\end{tabular}}
\label{tab4}
\end{table*}

\textbf{\textit{Stanford Drone Dataset}}\cite{robicquet2016learning}\footnote{\url{http://cvgl.stanford.edu/projects/uav_data/}}: The Stanford Drone Dataset is a large-scale object tracking dataset, that was made public by Stanford University in 2016. These video sequences were captured in a real campus environment by a 4k camera on a quadcopter, which hovered above various intersections on campus with a flight height of about 80 meters. This dataset contains 10 object types with more than 19,000 objects, including 112,000 pedestrians, 64,000 bicycles, 13,000 cars, 33,000 skateboarders, 22,000 golf carts, and 11,000 public cars, all of which can be used for multiple object tracking. Although this dataset only has videos of a college campus, the data has enough pluralism to be applied in various scenarios.

\textbf{\textit{UAV123 Dataset}}\cite{mueller2016benchmark}\footnote{\url{https://cemse.kaust.edu.sa/ivul/uav123}}: The UAV123 dataset is a long-term aerial object tracking dataset, which was designated as public by King Abdullah University of Science and Technology in 2016. It contains 123 video sequences and more than 110,000 representative frames. The label information of each sequence adopts a horizontal bounding box (i.e., upper left and lower right), and the bounding box size and aspect ratio show significant differences from the first frame. These video sequences were captured by three different UAVs: an off-the-shelf professional-grade UAV (DJIS1000) with a flight height of 5-25 meters, a small low-cost UAV, and a UAV simulator. The UAV123 dataset has multiple variations of scenes, objects, and their corresponding attitudes, making it better suited for a deep learning framework.

\textbf{\textit{Drone Tracking Benchmark (DTB70)}}\cite{li2017visual}\footnote{\url{https://link.zhihu.com/?target=https\%3A//github.com/flyers/drone-tracking}}: The DTB70 dataset includes both short-term and long-term aerial objects, which were provided by the Hong Kong University of Science and Technology in 2017. It contains 70 video sequences. Some of these video sequences were captured in a real outdoor environment by a DJI Phantom 2 Vision+ drone, which hovered over the university campus with a flight altitude of lower than 120 meters. The others were intercepted through YouTube to increase the diversity of samples. Each frame contains $1280\times 720$ and its label information adopts a horizontal bounding box (i.e., upper left and lower right).

\textbf{\textit{Car Parking Lot Dataset (CARPK)}}\cite{Hsieh2017ICCV}\footnote{\url{https://lafi.github.io/LPN/}}: The CARPK dataset is a large-scale vehicle detection and counting dataset, which was designated as public by the National Taiwan University in 2017. In particular, it is the first and largest parking lot dataset acquired by drone views and is used for vehicle counting parked in a different parking lot. The dataset was acquired by a Phantom 3 Professional drone with a flight height of 40 meters, covering nearly 90,000 cars in four different parking lots. The maximum size of vehicles in the CARPK dataset is much larger than $64\times 64$, and the maximum number of cars in a single scenario in the CARPK dataset is 188. The label information of each vehicle adopts a horizontal bounding box (i.e., upper left and lower right). 

\textbf{\textit{Okutama-Action Dataset}}\cite{barekatain2017okutama}\footnote{\url{http://okutama-action.org/}}: Okutama dataset is a large-scale human action detection dataset, which was designated as public by five universities, including Munich University of Technology and the Royal Institute of Technology of Sweden, in 2017. It contains 43 video sequences with 77,365 representative frames. These video sequences were captured at 45- or 90-degree camera angles using two drones with a flight height of 10-45 meters. In addition, the position and orientation of the UAV are flexible and changeable in order to acquire the diversification of the object. This dataset covers 12 action types, such as reading, handshaking, drinking, and carrying. The speed of recorded videos is 30 frames per second (fps), and the image size is $3840\times 2160$.

\textbf{\textit{UAV Detection and Tracking (UAVDT) Dataset}}\cite{du2018unmanned}\footnote{\url{https://sites.google.com/site/daviddo0323/projects/uavdt}}: 
The UAVDT dataset is a large-scale vehicle detection and tracking dataset, which was designated as public by the University of the Chinese Academy of Sciences in 2018. It contains 100 video sequences with 80,000 representative frames, approximately 2,700 vehicles with 0.84 million bounding boxes, covering a range of weather conditions, occlusion, and flying heights. This dataset presents all sorts of common scenarios, including squares, arterial roads, toll stations, highways, intersections, and T-junctions. The speed of recorded videos is 30 frames per second (fps), and the image size is $1080\times 540$ pixels, which can be used for multiple tasks, such as vehicle detection, single vehicle tracking, and multiple vehicle tracking.

\textbf{\textit{DAC-SDC dataset}\cite{xu2019dac}\footnote{\url{www.github.com/xyzxinyizhang/2018-DAC-System-Design-Contest}}: 
The Design Automation Conference (DAC) is a challenging object detection dataset collected by UAV, which was designated as public by the University of Notre Dame in 2018. It contains 95 categories and 150k images captured with different points of UAV view. Each extracted frame includes $640\times 360$ pixels.}

\textbf{\textit{VisDrone2018 Dataset}}\cite{zhu2018vision}\footnote{\url{https://github.com/VisDrone/VisDrone-Dataset}}: The VisDrone2018 dataset is a large-scale visual object detection and tracking dataset, which was designated as public by three universities, Tianjin University, GE Global Research, and Temple University, in 2018. It contains 263 video sequences with 179,264 representative frames and 10,209 static images. These video sequences were captured by various camera devices using multiple drones (i.e., DJI Mavic and Phantom series (3, 3A, 3SE, 3P, 4, 4A, 4P)), which hovered above 14 cities in China. This dataset covers multiple common objects, such as pedestrians, cars, bicycles, and tricycles. The maximum image size of each video are much larger than $2000\times 1500$, and they can be used for multiple tasks, particularly object detection, single object tracking, and multiple object tracking. There are over 2.5 million objects with their label information in a horizontal bounding box. 

\textbf{\textit{VisDrone2019 Dataset}}\cite{zhu2020vision}\footnote{\url{https://github.com/VisDrone/VisDrone-Dataset}}: Compared to VisDrone2018, VisDrone2019 added 25 long-term tracking video sequences with a total of 82,644 frames, of which 12 clips were acquired in the daytime, and the rest were by night. Therefore, this dataset contains 288 video sequences with 261,908 representative frames and 10,209 static images. For each target, the scaling is much smaller and the disturbance factor is much greater.

\textbf{\textit{Moving Object Recognition (MOR-UAV) Dataset}}\cite{mandal2020mor}\footnote{\url{https://arxiv.org/abs/2008.01699}}: 
The MOR-UAV dataset is a large-scale video dataset for moving object recognition in UAV videos, which was designated as public by the Malaviya National Institute of Technology Jaipur in 2020. It contains 30 video sequences with 10,948 representative frames, and approximately 89,783 moving object instances, covering various challenging scenarios such as night time, occlusion, camera motion, weather conditions, camera views, and so on. MOR-UAV can be used as the benchmark for both MOR and moving object detection (MOD) in UAV videos. The videos are recorded at 30 frames per second (fps) and the image size varies from $1280\times 720$ to $1920\times 1080$ pixels. The moving objects are labeled using the Yolo-mark1 tool, and about 10,948 frames are annotated representing moving vehicles. There are two categories of vehicles: cars and heavy vehicles.

\textbf{\textit{DroneVehicle dataset}}\cite{zhu2020drone}\footnote{\url{https://github.com/VisDrone/DroneVehicle}}: The DroneVehicle dataset is large-scale object detection and counting dataset with both RGB and thermal infrared (RGBT) images captured by camera-equipped drones, which was designated as public by the Tianjin University in 2020. It contains 15,532 pairs of images, i.e., RGB and infrared images, covering challenging scenarios with illumination, occlusion, and scale variations. DroneVehicle dataset can be used as the benchmark for both object detection and counting on the UAV platform. The images in this dataset were captured over various urban areas, including urban roads, residential areas, parking lots, highways, etc., from day to night. The image size is $840\times 712$ pixels.

\textbf{\textit{AU-AIR dataset}}\cite{bozcan2020air}\footnote{\url{https://bozcani.github.io/auairdataset}}: 
The multi-purpose aerial dataset (AU-AIR) is a large-scale object detection dataset from multi-modal sensors (i.e., visual, time, location, altitude, IMU, velocity) captured by camera-equipped drones, which was designated as public by the Aarhus University in 2020. It contains 8 video sequences with 32,823 extracted frames at the intersection of Skejby Nordlandsvej and P.O Pedersensvej (Aarhus, Denmark) on windless days with various lighting and weather conditions. This dataset contains 8 object types, including person, car, bus, van, truck, bike, motorbike, and trailer, all of which can be used for static or video object detection. Each frame contains $1920\times 1080$ pixels.

\textbf{\textit{BIRDSAI dataset}}\cite{bondi2020birdsai}\footnote{\url{https://sites.google.com/view/elizabethbondi/dataset}}: The benchmarking IR dataset for surveillance with aerial intelligence (BIRDSAI) is a challenging object detection and tracking dataset collected using a TIR camera mounted on a fixed-wing UAV in multiple
African protected areas, which was designated as public by the Harvard University in 2020. It contains 48 real aerial TIR videos of varying lengths and 124 synthetic aerial TIR videos generated from AirSim-W. This dataset contains humans and animals with scale variations, background clutter, large camera rotations, and motion blur, etc. Each frame contains $640\times 480$ pixels.

\textbf{\textit{MOHR dataset}}\cite{zhang2021empirical}: 
The benchmarking IR dataset is a large-scale benchmark object detection dataset collected at different altitudes by employing three cameras, i.e., DJI Phantom 4Pro, Sonny RX1rM2, and Nikon D800. The dataset includes 3,048 images of size $5482\times 3078$, 5,192 images of size $7360\times 4912$, and 2,390 images of size $8688\times 5792$, respectively. It contains 90,014 object instances with labels and bounding boxes were annotated, which includes 25,575 cars, 12,957 trucks, 41,468 buildings, 7,718 flood damages, and 2,296 collapses, covering the challenging of scale variations.

\textbf{\textit{UVSD dataset}}\cite{zhang2020multi}\footnote{\url{https://github.com/liuchunsense/UVSD}}: The UAV-based vehicle segmentation dataset (UVSD) is a large-scale benchmark object detection, counting, and segmentation dataset. The dataset includes 5,874 images, with 98,600 object instances with high-quality instance-level semantic annotations. These images are captured by DJI matrice 200 quadcopter integrated with a zenmuse X5S gimbal and camera, and image size varies from $960\times 540$ to $5280\times 2970$ pixels. In particular, UVSD has multiple format annotations, including pixel-level semantic, OBB and HBB.

\begin{figure*}[!t]
    \centering\includegraphics[width=1.0\textwidth]{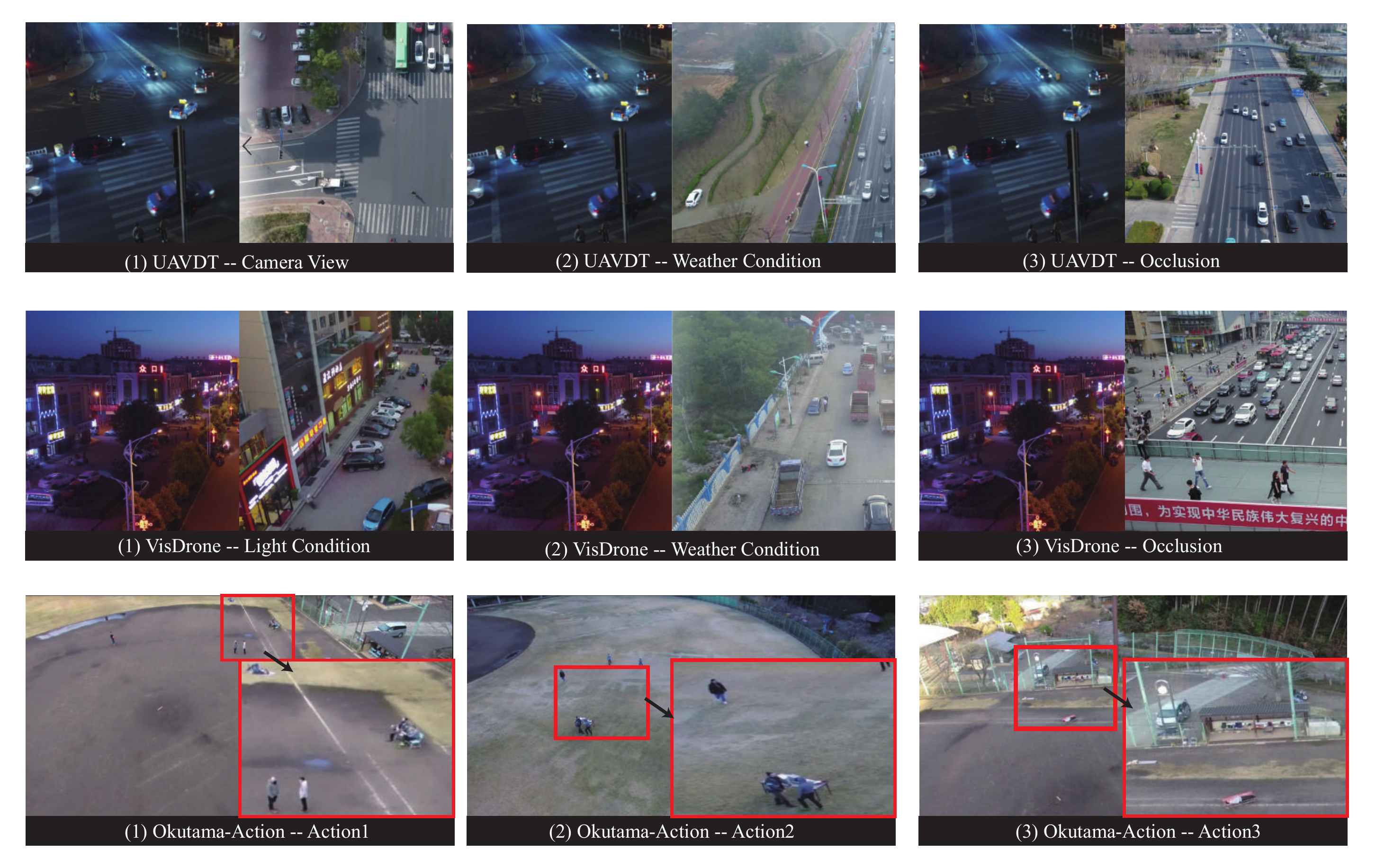}
    \caption{Visual samples of annotated images taken from benchmark datasets. The first, second, and third rows stand for UAVDT, VisDrone, and Okutama-Action datasets, respectively.}
\label{fig:example}
\end{figure*}

\section{Experiment Results And Analysis} \label{sec:Exp}
In this section, we take four benchmark datasets, including VisDrone, UAVDT, Okutama-Action and Stanford UAV datasets, to illustrate the performance of representative object detection and tracking methods. Fig. \ref{fig:example} shows examples of annotated images in these four datasets.

\subsection{Evaluation of Object Detection from UAV-borne Images} \label{sec:Exp_DET}
For object detection from UAV-borne images, the common performance metrics are Average Precision (AP) and Average Recall (AR). AP is used as a global measure. 
More precisely, the value of AP and AR are related to the rate between the overlap of the detection bounding box and the ground-truth box exceeds a certain percentages. The most frequently used is $AP^{IoU=0.50:0.05:0.95}$, $AP^{IoU =0.50}$, $AP^{IoU=0.75}$, $AR^{max=1}$, $AR^{max=10}$, $AR^{max=100}$ and $AR^{max =500}$. Specifically, $AP^{IoU=0.50:0.05:0.95}$ denotes the mean average precision (mAP), that is, the average value of the multiple intersection over union (IOU) threshold, which is defined as the geometric overlap between predictions and ground truths, of all categories with step size of 0.05. $AP^{IoU=0.50}$ and $AP^{IoU=0.75}$ are computed at a certain IOU threshold over all categories. Moreover, $AP^{s}=AP^{small}$, $AP^{m}=AP^{medium}$, $AP^{l}=AP^{large}$ represent the average precision at different scales. $AR^{max=1}$, $AR^{max=10}$, $AR^{max=100}$ , and $AR^{max=500}$ are the maximum recalls number of 1, 10, 100, and 500 detected objects in each image. For more details please refer to \cite{zhu2018vision, zhu2020vision}.

Table \ref{tab:ExpDet} lists quantitative results of several state-of-the-art detection methods. Their experiment results are distributed in different UAV object detection datasets and most of them just use $AP=AP^{IoU=0.50:0.05:0.95}$ as the only evaluation criteria. To be fair, the performance of these works is compared according to their AP value under a specific dataset.

\textbf{VisDrone dataset:} This dataset has severe sample imbalance and occlusion problems between small objects. NDFT with domain-robust features, which transfers the learned NDFT through UAVDT to VisDrone dataset, achieves the best performance among all comparative methods, i.e., 52.77\% AP score on the VisDrone-DET validation set due to the testing set has been closed after the ICCV2019 conference. The possible reason is that NDFT could achieve a substantial gain in robustness to many UAV-specific nuisances, such as varying flying altitudes, adverse weather conditions, dynamically changing viewing angles, etc. SAMFR with spatial-refinement module and receptive field expansion block (RFEB), MPFPN with parallel branch becomes the second and third with 33.72\% and 29.05\% AP scores.

Table \ref{tab:ExpDet} shows the results of 10 baseline methods in the VisDrone-DET2019 Challenge, i.e.,  FPN \cite{lin2017feature}, R-FCN \cite{girshick2015fast}, Faster R-CNN (FRCNN) \cite{ren2015faster}, SSD \cite{liu2016ssd}, Cascade CNN \cite{cai2018cascade}, RetinaNet \cite{lin2017focal}, CornetNet \cite{law2018cornernet}, RefineNet \cite{zhang2018single}, DetNet \cite{li2018detnet}, and Light Faster R-CNN (Light-RCNN) \cite{li2017light}. The samples are in strict accordance, with 6,471 for training, 548 for validation and 1,580 for testing. For the parameters of these networks, we adjust them within a reasonable range or directly adopted the default values. CornerNet achieves the best performance, while SSD$^*$ performs the worst.

\textbf{UAVDT dataset:} With different locations but similar environments to the VisDrone dataset, UAVDT has higher complexity due to its images collected from a variety of scenes. Moreover, the weather condition would increase the difficulty of single, multiple, or overlapping small object detection. D2Det published in CVPR2020 with dense local regression achieves the best performance among all methods, i.e., 56.92\% AP score on the testing set. NDFT with domain-robust  features, FPN employed ResNet101 becomes the second and third with 52.03\% and 49.05\% AP scores. We also report the detection results of 8 baseline DL-based networks, including R-FCN \cite{girshick2015fast}, Faster R-CNN (FRCNN) \cite{ren2015faster}, FRCNN plus FPN \cite{lin2017feature}, SSD \cite{liu2016ssd}, Cascade CNN \cite{cai2018cascade}, Reverse connection with Objectness prior Networks (RON) \cite{lin2017focal}, ClusDet \cite{yang2019clustered}, and DMDet \cite{li2020density}, are shown in Table \ref{tab:UAVDTDet}. Among them, the image size for UAVDT was $1024\times 540$ pixels, while the sample size of some methods varied. The network parameters were the same as the VisDrone dataset. FPN achieves the best performance, while RON performs the worst.

\begin{table*}[!t]
\caption{Performance comparisons of UAV exclusive detection networks and classic detection networks. The best performers are highlighted in bold.}
\small
\centering
\resizebox{1\textwidth}{!}{
\begin{tabular}{l||l|c|c|c|c|c|c|c|c|c|c}
\toprule[1.5pt]
Method& Network & Train/Test & Image Size &AP & AP\_50 & AP\_75 & AR\_1 &AR\_10 & AR\_100 & AR\_500 & Exp. Data\\ 
\hline \hline
Yang et al\cite{yang2019effective} & VGG-16 & 3,475/869 & - & \textbf{92.00} & - & - & - & - & -  & - & Own vehicle \\
UAV-YOLO\cite{liu2020uav} & YOLOv3 & 3,776/630 & $608 \times 608$ & \textbf{90.86} & - & - & - & - & - & - & UAV123+Own \\ 
FS-SSD\cite{liang2019small} & VGG16 & 989/459 & $512 \times 512$ & \textbf{89.52} & - & - & - & - & - & - & CARPK \\ \hline
FS-SSD\cite{liang2019small} & VGG16 & 69,673/53,224 & $512 \times 512$ & 65.84 & & & & & & & Stanford Drone \\ 
Yang et al\cite{yang2019effective} & VGG-16 & 3,500/831 & $320 \times320$ & \textbf{90.40} & - & - & - & - &  - & - & Stanford Drone \\ \hline
MSOA-Net\cite{MSOA-Net2020} & ResNet50 & 3,564/1,725 & $1333 \times800$ & 77.00 & \textbf{91.50} & \textbf{83.30} & & & & & UAVDT\\
GDF-Net\cite{GDF-Net2020} & ResNet50 & 11,915/16,580 & $1200 \times675$ & 15.40 & 26.10 & 17.00 & \textbf{13.20} & \textbf{23.10} & \textbf{27.60} & & UAVDT\\
DSYolov3\cite{DSYolov32021} & Yolov3 & 24,143/16,592 & $1200 \times 540$ & 9.80 & 23.40 & 5.00 & - & - & - & & UAVDT \\
ClusDet\cite{ClusDet2019} & ResNeXt101 & 23,238/15,069 & $1080 \times540$ & 13.70 & 25.50 & 12.50 & - & - & - & - & UAVDT\\ 
DSHNet\cite{yu2021towards} & ResNet50 & 23,258/15,069 & $1080 \times 540$ & 17.80 & 30.40 & 19.70 & - & - & - & - & UAVDT \\
D2Det\cite{cao2020d2det} & ResNet101 & 23,258/15,069 & $1,333 \times 800$ & 56.92 & - & - & - & - & - & - & UAVDT \\ 
NDFT\cite{wu2019delving} & ResNet101 & 23,258/15,069 & $1,333 \times 800$ & 52.03 & - & - & - & - & - & - & UAVDT\\ 
DSHNet\cite{yu2021towards} & ResNet50 & 23,258/15,069 & $1,080 \times 540$ & 17.80 & 30.4 & 19.7 & - & - & - & - & UAVDT\\ 
DNOD\cite{TIAN2021292} & YOLOv4 & 23,258/15,069 & $1080 \times 540$ & 14.20 & 31.90 & 11.00 &  &- &-&- & UAVDT \\
DNOD\cite{TIAN2021292}  & EfficientDet-D7& 23,258/15,069 & $1080 \times 540$ & 12.90 & 32.00	& 10.90 & - &- &-&- &UAVDT\\
FPN$^*$\cite{lin2017feature}& FPN &23,258/15,069 & $1080\times 540$&49.05&- & &- &- &- &- & UAVDT\\ 
RON\cite{kong2017ron}   &VGG16     &23,258/15,069  & $1080 \times 540$  &5.0 &15.9 &1.7 &- &- &-&-&UAVDT \\
RetinaNet\cite{lin2017focal} &RetinaNet &23,258/15,069& $1080 \times 540$  &33.95 &- & -&- &- &- & - & UAVDT\\ \hline
ECas\_RCNN\cite{ECascade-RCNN} & ResNet50 & 6,371/521 & $1450 \times800$ & 28.40 & - & - & - & - & - & - & VisDrone-Val \\ 
 
GDF-Net\cite{GDF-Net2020} & ResNet50 & 6,471/11,610 & $1200 \times675$ & 18.20 & 30.80 & 19.20 & 8.10 & 24.10 & 28.70 & & VisDrone-Val\\

HRDNet\cite{liu2020HRDNet} & ResNeXt50+101 & 3,564/1,725 & $3800 \times3800$ &  35.51 & \textbf{62.00} & \textbf{35.13} & 0.39 & 3.38 & 30.91 & \textbf{46.62} & VisDrone-Val\\ 
D-A-FS SSD\cite{ICIT2020} & - & - &- & - & - & - & - & - & - & - & VisDrone-Val\\ 
ClusDet\cite{ClusDet2019} & ResNeXt101 & 6,471/548 & $2000 \times1500$ & 32.40 & 56.20 & 31.60 & - & - & - & - & VisDrone-Val\\ 
CenterNet\cite{pailla2019object} & HourGlass-104 & 3,564/1,725 & $1024 \times1024$ & 21.58 & 48.09 & 16.76 & \textbf{12.04} & \textbf{29.60} & \textbf{39.63} & 40.42 & VisDrone-Val \\ 
DSHNet\cite{yu2021towards} & ResNet50 & 6,471/548 & $2000 \times1500$ & 30.30 & 51.80 & 30.90 & - & - & - & - & VisDrone-Val \\
NDFT\cite{wu2019delving} & ResNet101 & 6,471/548 & $2,000 \times 1,500$ & 52.77 & - & - & - & - & - & - & VisDrone-Val\\
DSHNet\cite{yu2021towards} & ResNet50 & 6,471/548 & $2,000 \times 1,500$ & 24.60 & 44.40 & 24.10 & - & - & - & - & VisDrone-Val \\ 
MPFPN\cite{liu2020small} & ResNet101 & 6,471/1,580 & $1440\times 800$ & 29.05 & 54.38 & 26.99 & 0.55 & 5.81 & 35.57 & 45.69 & VisDrone-Val\\
SAMFR\cite{wang2019spatial} & DetNet59 & 6,471/548 & $512 \times 512$ & 33.72 & \textbf{58.62} & 33.88 & 0.53 & 3.40 & 22.60 & 46.03 & VisDrone-Val \\
DANN\cite{jadhav2020aerial} & RetinaNet & 6,471/548 & $1500 \times 1000$ & 11.19 & 25.65 & 8.78 & 0.56 & 4.87 & 17.19 & 24.09 & VisDrone-Val \\
Cas\_RCNN+FPN\cite{youssef2021automatic} & ResNet101 & 4,960/1,534 & $1500 \times 2000$ & 20.46 & 38.58 & 18.83 & 1.32 & 11.32 & 25.82 & 25.84 & VisDrone-Val \\  
DNOD\cite{TIAN2021292} & YOLOv4 & 6,471/1,610 & $1260 \times 765$ & 54.88 & - & - & - & - & - & - & VisDrone-Val \\ 
DNOD\cite{TIAN2021292} & EfficientDet-D7 & 6,471/1,610 & $1260 \times 765$ & 53.76 & - & - & - & - & - & - & VisDrone-Val \\ \hline
RRNet\cite{chen2019rrnet}  & HourGlass & 6,741/1580 & $512 \times512$ & \textbf{29.13} & \textbf{55.82} & \textbf{27.23} & \textbf{1.02} & \textbf{8.50} & \textbf{35.19} & \textbf{46.05} & VisDrone-Det\\
DSYolov3\cite{DSYolov32021} & Yolov3 & 6,471/548 & $1920\times 1080$ & 22.30 & 44.50 & 20.30 & - & - & - & - & VisDrone-Det \\
SAMFR\cite{wang2019spatial}& DetNet59& 6,471/1,580 & $512 \times 512$ & 20.18 & 40.03 & 18.42 & 0.46 & 3.49 & 21.6 & 30.82 & VisDrone-Det  \\ 
SyNet\cite{albaba2020synet} & CenterNet & 6,471/1,580 & $2000 \times 1500$ & 25.10 & 48.40 & 26.20 & - & - & - & - & VisDrone-Det \\ 
SlimYOLOv3\cite{zhang2019slimyolov3} & YOLOv3-SPP3-90 & 6,471/548 & $832 \times 832$ & 23.90 & - & - & - & - & - & - & VisDrone-Det \\ 
Zhang et al\cite{zhang2019dense} & ResNet50+RPN & 6,471/1580 & $2000 \times 1500$ & 22.61 & 45.16 & 19.94 & 0.42 & 2.84 & 17.1 & 35.27 & VisDrone-Det \\ 
CornerNet$^*$\cite{law2018cornernet} &CornetNet  &6,471/1,580& $2000 \times 1500$ &17.41 &34.12 &15.78 &0.39 &3.32 &24.37 &26.11 & VisDrone-Det\\
FPN$^*$\cite{lin2017feature}&FPN&6,471/1,580 & $2000 \times 1500$&16.51 &32.20 &14.91 &0.33 &3.03 &20.72 &24.93 & VisDrone-Det\\ 
Light-RCNN$^*$\cite{li2017light} &Light-RCNN &6,471/1580 & $2000 \times 1500$&16.53 &32.78 &15.13 &0.35 &3.16 &23.09 &25.07 & VisDrone-Det\\
Cas\_RCNN$^*$\cite{cai2018cascade} & Cascade R-CNN&6,471/1,580& $2000 \times 1500$ &16.09 &31.91 &15.01 &0.28 &2.79 &21.37 &28.43 & VisDrone-Det\\
DetNet59$^*$\cite{li2018detnet} &DetNet-59 &6,471/1,580& $2000 \times 1500$ &15.26 &29.23 &14.34 &0.26 &2.57 &20.87 &22.28 & VisDrone-Det\\
RefineNet\cite{zhang2018single} &RefineNet &6,471/1,580& $2000 \times 1500$  &14.90 &28.76 &14.08 &0.24 &2.41 &18.13 &25.69  & VisDrone-Det\\ 

RetinaNet$^*$\cite{lin2017focal}&RetinaNet&6,471/1,580& $2000 \times 1500$  &11.81 &21.37 &11.62 &0.21 &1.21 &5.31 &19.29  & VisDrone-Det\\
R-FCN$^*$\cite{girshick2015fast}   &  R-FCN    &6,471/1,580& $2000 \times 1500$   &7.20 &15.17 &6.38 &0.88 &5.35 &12.04 &13.95  & VisDrone-Det\\ 
FRCNN$^*$\cite{ren2015faster} &FRCNN      &6,471/1,580& $2000 \times 1500$   &3.55 &8.75 &2.43 &0.66 &3.49 &6.51 &6.53 & VisDrone-Det\\ 
SSD$^*$\cite{liu2016ssd}   &SSD  &6,471/1,580& $2000 \times 1500$  &2.52 &4.78 &2.47 &0.58 &2.81 &4.51 &6.41 & VisDrone-Det\\
\bottomrule[1.5pt]
\end{tabular}}
\label{tab:ExpDet}
\end{table*}

\begin{table*}[!t]
\caption{Object detection results on the UAVDT-DET testing set. The best performers are highlighted in bold.}
\tiny
\small
\centering
\resizebox{1\textwidth}{!}{
\begin{tabular}{l||l|c|c|c|c|c|c|c|c|c}
\toprule[1.5pt]
Method& Backbone & Train/Test & Image Size &AP & AP\_50 & AP\_75 & AP\_s &AP\_m & AP\_l & Exp. Data\\ 
\hline \hline
R-FCN \cite{girshick2015fast} & ResNet50 &23,258/15,069  & $1080 \times 540$  &7.0 &17.5 &3.9 &4.4 &14.7 &12.1 & UAVDT\\
SSD \cite{liu2016ssd}   &VGG16      &23,258/15,069  & $1080 \times 540$  &9.3 &21.4 &6.7 &7.1 &17.1 &12.0  & UAVDT\\  
FRCNN \cite{ren2015faster} &VGG16      &23,258/15,069  & $1080 \times 540$  &5.8 &17.4 &2.5 &3.8 &12.3 &9.4 & UAVDT\\ 

FRCNN \cite{ren2015faster}+FPN \cite{lin2017feature} & ResNet50 &23,258/15,069 & $1080 \times 540$  &11.0 &23.4 &8.4 &8.1 &20.2 &26.5& UAVDT \\ 
ClusDet \cite{yang2019clustered} & ResNet50 &-/25,427 & $1080 \times 540$  &13.7 &\textbf{26.5} &12.5 &9.1 &25.1 &31.2 & UAVDT\\
DMDet \cite{li2020density} & ResNet50  &-/32,764 & $1080 \times 540$  &\textbf{14.7} & 24.6 & \textbf{16.3} &\textbf{9.3} &\textbf{26.2} &\textbf{35.2} & UAVDT\\
\bottomrule[1.5pt]
\end{tabular}}
\label{tab:UAVDTDet}
\end{table*}

\begin{table*}[!t]
\caption{Performance comparisons of UAV exclusive detection networks and classic detection networks for the VisDrone-VID testing set. The best performers are highlighted in bold.}
\centering
\resizebox{1\textwidth}{!}{
\begin{tabular}{l||l|c|c|c|c|c|c|c|c|c|c}
\toprule[1.5pt]
Method& Framework & Train/Test & Image Size &AP & AP\_50 & AP\_75 & AR\_1 &AR\_10 & AR\_100 & AR\_500 & Exp. Data\\ \hline \hline
TDFA\cite{xie2021two} & FlowNet+Fea\_Agg & 54,503/14,114 &$720\times 1280$ &- & 87.18 &- &- &- &- &- & Okutama\\
STDnet-ST\cite{bosquet2021stdnet} & STDnet+RCN & 23,829/16,580 & $1024\times 540$ & 13.30 & 36.40 &-&- &- &- &- & UAVDT\\ 
Zhang et al\cite{zhang2020drone}& Cas\_RCNN+IRR-PWC & 17,268/5,397 & $1280 \times 720$ & \textbf{65.20} &\textbf{88.80} &\textbf{74.60} & -&- &- &- & VisDrone-VID \\
STCA\cite{Pi_2019_ICCV}  & F-SSD+FCOS & 24,198/6,322 & -& 18.73 & 44.38 & 12.68 &- & -& -& -& VisDrone-VID\\ 
TDFA\cite{xie2021two} & FlowNet+Fea\_Agg & 24,201/2,819  &$720\times 1280$ & 27.27 & 50.73 & 27.94 &- & -&- &- & VisDrone-VID\\  
STDnet-ST\cite{bosquet2021stdnet}&STDnet+RCN & 24,201/6,635 & $1,920\times 1,080$ & 7.50 & 22.40 &- &- &- &- & -& VisDrone-VID\\  
FGFA$^*$ \cite{zhu2017flow,zhu2019visdrone} &VGG16 &24,198/6,322 & 3840$\times$ 2160&18.33 &39.71 &14.39 &10.09 &26.25 &34.49 &34.89 & VisDrone-VID\\ 
CFE-SSDv2 \cite{zhao2018cfenet,zhu2019visdrone} &SSD &24,198/6,322 & 3840$\times$ 2160&21.57 &44.75 &17.95 &\textbf{11.85} &\textbf{30.46} &\textbf{41.89} &\textbf{44.82} & VisDrone-VID\\
D\&T (R-FCN) \cite{feichtenhofer2017detect,zhu2019visdrone} &Hourglass  &24,198/6,322 & 3840$\times$ 2160&17.04 &35.37 &14.11 &10.47 &25.76 &31.86 &32.03 & VisDrone-VID\\
FPN$^*$ \cite{lin2017feature,zhu2019visdrone} &ResNet-101 &24,198/6,322 & 3840$\times$ 2160&16.72 &39.12 &11.80 &5.56 &20.48 &28.42 &28.42 & VisDrone-VID\\ 
CornerNet$^*$ \cite{law2018cornernet,zhu2019visdrone} &Hourglass-59 &24,198/6,322 & 3840$\times$ 2160&16.49 &35.79 &12.89 &9.47 &24.07 &30.68 &30.68 & VisDrone-VID\\
CenterNet$^*$ \cite{zhou2019objects,zhu2019visdrone} &Hourglass &24,198/6,322 & 3840$\times$ 2160&15.75 &34.53 &12.10 &8.90 &22.80 &29.20 &29.20 & VisDrone-VID\\
Faster R-CNN$^*$ \cite{ren2015faster,zhu2019visdrone} &VGG16 &24,198/6,322 & 3840$\times$ 2160&14.46 &31.80 &11.20 &8.55 &21.31 &26.77 &26.77 & VisDrone-VID\\
RD \cite{zhang2018single,zhu2018visdrone}   & RefineDet&24,198/6,322 & 3840$\times$ 2160&14.95 &35.25 &10.11 &9.67 &24.60 &29.72 &29.91 & VisDrone-VID\\
RetinaNet\_s \cite{lin2017focal,zhu2018visdrone} &RetinaNet &24,198/6,322 & 3840$\times$ 2160 & 8.63 &21.83 &4.98 &5.80 &12.91 &15.15 &15.15 & VisDrone-VID\\
\bottomrule[1.5pt]
\end{tabular}}
\label{tab:ExpVID}
\end{table*}

\begin{table*}[!t]
\caption{Video object detection results on the Okutama-Action and UAVDT testing set. ``
\#vid'' is the number of videos that send to the detector. The best performers are highlighted in bold.}
\small 
\centering
\resizebox{1\textwidth}{!}{
\begin{tabular}{l|l|c|c|c|c||l|c|c|c|c|c}
\toprule[1.5pt]
Method& Backbone & \#vid &Image Size & AP\_50 & Exp.Data&Method& Train/Test & Image Size &AP & AP\_50 & Exp.Data\\ 
\hline \hline
SSD \cite{liu2016ssd}   &VGG      &10 &$960\times540$ &18.80 & Okutama-Action &Faster RCNN \cite{du2018unmanned} & 23,829/76,215 & $1080\times 540$ &6.6 & 26.00& UAVDT\\
SSD \cite{liu2016ssd}   &ResNet50      &10 &$608\times608$ &52.30 & Okutama-Action & SSD \cite{du2018unmanned}& 23,829/76,215 & $1080\times 540$  &6.0 & 23.50& UAVDT\\
R-FCN \cite{girshick2015fast}   &ResNet50      &10 &$608\times608$ &53.50 & Okutama-Action &R-FCN \cite{du2018unmanned}  & 23,829/76,215 & $1080\times 540$ &9.2 & \textbf{32.50}& UAVDT\\
Retinanet \cite{lin2017focal}   &ResNet50      &10 &$608\times608$ &\textbf{56.30} & Okutama-Action & FGFA \cite{zhu2017flow} & 23,829/76,215 & $1080\times 540$  &6.3 & 20.70& UAVDT\\
YOLOv3\_tiny \cite{adarsh2020yolo}   &DarkNet-53     &10 &$608\times608$ &52.40 & Okutama-Action & FPN\cite{lin2017feature} & 23,829/76,215 & $1080\times 540$ &\textbf{11.8} & 29.70& UAVDT\\
\bottomrule[1.5pt]
\end{tabular}}
\label{tab:Okutama/UAVDTVid}
\end{table*}

\begin{table*}[!t]
\caption{Performance comparisons of UAV exclusive tracking networks and classic tracking networks for the VisDrone-MOT testing set taken MOTA, MOTP, etc as evaluation indexes. The best performers are highlighted in bold.}
\small 
\centering
\resizebox{1\textwidth}{!}{
\begin{tabular}{l||l|c|c|c|c|c|c|c|c|c|c|c|c|c}
\toprule[1.5pt]
\textbf{Method}& \textbf{Framework} & \textbf{Train/Test(seq)} & \textbf{Image Size} &\textbf{MOTA} & \textbf{MOTP}& \textbf{IDF1} &\textbf{FAF} &\textbf{MT} &\textbf{ML} & \textbf{FP}  & \textbf{FN}& \textbf{IDS} & \textbf{FM} & \textbf{Exp. Data}\\ 
\hline \hline
\multirow{2}{*}{TNT\cite{zhang2019eye}} &RetinaNet50  &\multirow{2}{*}{56/33} & \multirow{2}{*}{$3840\times 2160$} &\multirow{2}{*}{\textbf{48.6}} & \multirow{2}{*}{-}&\multirow{2}{*}{\textbf{58.1}} &\multirow{2}{*}{-} &\multirow{2}{*}{281} &\multirow{2}{*}{478} &\multirow{2}{*}{\textbf{5,349}} & \multirow{2}{*}{76,402} & \multirow{2}{*}{468} & \multirow{2}{*}{-}& \multirow{2}{*}{VisDrone-MOT}\\
&+TrackletNet  && & & && &&&&&  \\ 
HDHNet\cite{huang2021multiple} & HRNet+DLA & 56/7 & $3840\times 2160$ & 32.9 & \textbf{76.9} & 42.3 &- & -& -& 80,454 & \textbf{35,686} & 1,056 & 1,242 & VisDrone-MOT\\ 
Flow-Tracker\cite{li2019multiple} & IOU+Optical flow & 56/7 & $3840\times 2160$ & 26.4 & 78.1 & 41.9 &- & 115 & \textbf{246} & 9,987 & 43,766 & \textbf{127} & \textbf{428} & VisDrone-MOT\\ 
CMOT$^*$ \cite{bae2014robust} &Faster RCNN & 56/16& $3840\times 2160$ &31.5 & 73.3& 51.3 &1.42 &282 &435 & 26,851 & 72,382 &789 & 2,257 &  VisDrone-MOT\\ 
TBD$^*$ \cite{geiger20133d}  &Faster RCNN &56/16& $3840\times 2160$ &35.6 & 74.1& 45.9 &1.17 &302 &419 & 22,086 & 70,083 &1,834 & 2,307 &  VisDrone-MOT\\ 
$H^2T^*$ \cite{wen2014multiple} &Faster RCNN &56/16& $3840\times 2160$ &32.2 & 73.3& 44.4 &0.95 &214 &494 & 17,889 & 79,801 &1,269 & 2,035 &  VisDrone-MOT\\ 
IHTLS$^*$ \cite{dicle2013way}  &Faster RCNN &56/16& $3840\times 2160$ &36.5 & 74.8& 43.0 &0.94 &245 &446 & 14,564 & 75,361 &1,435 & 2,662 &  VisDrone-MOT\\ 
Ctrack \cite{al2017robust}  &Faster RCNN &56/16& $3840\times 2160$ &30.8 & 73.5&51.9 &1.95 &\textbf{369} &375 & 36,930 & 62,819 &1,376 & 2,190 &  VisDrone-MOT\\ 
CEM$^*$ \cite{milan2013continuous}  &Faster RCNN   &56/16 &$3840\times 2160$ &5.1 & 72.3& 19.2 &1.12 &105 &752 & 21,180 & 116,363 &1,002 & 1,858& VisDrone-MOT  \\
GOG$^*$ \cite{pirsiavash2011globally} &Faster RCNN    &56/16 &$3840\times 2160$ &38.4 & 75.1& 45.1 &\textbf{0.54} &244 &496 & 10,179 & 78,724 &1,114 & 2,012 & VisDrone-MOT  \\ 
\bottomrule[1.5pt]
\end{tabular}}
\label{tab:ExpMOT1}
\end{table*}

\begin{table*}[!t]
\caption{Performance comparisons of UAV exclusive tracking networks and classic tracking networks for the VisDrone-MOT testing set taken AP as evaluation indexes. The best performers are highlighted in bold.}
\small 
\centering
\resizebox{1\textwidth}{!}{
\begin{tabular}{l||l|c|c|c|c|c|c|c|c|c|c|c|c}
\toprule[1.5pt]
\textbf{Method}& \textbf{Framework} & \textbf{Train/Test(seq)} & \textbf{Image Size} &\textbf{AP} & \textbf{AP\_0.25} & \textbf{AP\_0.5} &\textbf{AP\_0.75} &\textbf{AP\_car} &\textbf{AP\_bus}& \textbf{AP\_truck}  &\textbf{AP\_ped}& \textbf{AP\_van} & \textbf{Exp. Data}\\ 
\hline \hline
PAS Tracker\cite{stadler2020pas} & CenterNet+IOU & 56/7 & $608\times 608$ & \textbf{50.80} & \textbf{66.10} & \textbf{52.50} & \textbf{33.80} & \textbf{62.7} & \textbf{81.20} & \textbf{43.90} & \textbf{30.30} & \textbf{35.90} & VisDrone-MOT\\
HMTT\cite{pan2019multi} & CenterNet+IOU & 56/7 & $608\times 608$ & 28.67 & 39.05 & 27.88 & 19.08 & 44.35 & 30.56 & 18.75 & 26.49 & 23.19 & VisDrone-MOT\\ 
DAN\cite{jadhav2020aerial} & RetinaNet+DAN & - & $1500\times 1000$ & 13.88 & 23.19 & 12.81 & 5.64 & 32.20 & 8.83 & 6.61 & 18.61 & 3.16 & VisDrone-MOT\\ 
GGD\cite{ardo2019multi}  & Faster RCNN & 56/33 & $3840\times 2160$ & 23.09 & 31.01 & 22.70 & 15.55 & 35.45 & 28.57 & 11.90 & 17.20 & 22.34 & VisDrone-MOT\\ 
Cas\_RCNN+FPNCas\_RCNN+FPN\cite{youssef2021automatic} & Cascade R-CNN & - & $2000\times 1500$ & 28.51 & 44.76 & 30.38 & 10.40 & 35.09 & 34.58 & 18.20 & - & 26.18 & VisDrone-MOT\\ 
\bottomrule[1.5pt]
\end{tabular}}
\label{tab:ExpMOT2}
\end{table*}

\begin{table*}[!t]
\caption{Performance comparisons of UAV exclusive tracking networks and classic tracking networks for the Stanford Droned dataset taken MOTA, MOTP, etc as evaluation indexes. The best performers are highlighted in bold.}
\small 
\centering
\resizebox{1\textwidth}{!}{
\begin{tabular}{l||l|c|c|c|c|c|c|c|c|c|c|c|c|c}
\toprule[1.5pt]
\textbf{Method}& \textbf{Framework} & \textbf{Train/Test(seq)} & \textbf{Image Size} &\textbf{MOTA} & \textbf{MOTP}& \textbf{IDF1} &\textbf{IDP} &\textbf{MT\%} &\textbf{ML\%} & \textbf{FP}  & \textbf{FN}& \textbf{IDS} & \textbf{FM} & \textbf{Exp. Data}\\ 
\hline \hline
\multirow{2}{*}{IPGAT\cite{yu2020conditional}} & LSTM+ CGAN & \multirow{2}{*}{36/24} & \multirow{2}{*}{$1080\times 540$} &\multirow{2}{*}{99.9} & \multirow{2}{*}{99.9}  &\multirow{2}{*}{90.0} & \multirow{2}{*}{90.0} & \multirow{2}{*}{99.8} & \multirow{2}{*}{\textbf{0.13}} & \multirow{2}{*}{3} & \multirow{2}{*}{\textbf{833}}  & \multirow{2}{*}{3, 395} &\multirow{2}{*}{\textbf{905}} & \multirow{2}{*}{Stanford Drone}\\
& +Siamese & &  & & & & & & & && & &  \\  
CEM \cite{milan2013continuous}  &Faster RCNN    &36/24 & $1417\times 2019$ & 3.0 &81.8  & 5.4 &47.6 & 2.7 & 90.25 &  972,646& 348,495 &3,103& 5,997 & Stanford Drone  \\ 
GOG \cite{pirsiavash2011globally}  &Faster RCNN  &36/24 & $1417\times 2019$ & 98.9 &\textbf{100.0}  & 86.3 &86.7& \textbf{100.0} & 96.2 &  3 &66,625& 4,928 &2,621 & Stanford Drone  \\
IOUT \cite{bochinski2017high} &Faster RCNN    &36/24 &$1417\times 2019$ & \textbf{99.9} &\textbf{100.0}  & 93.2 &93.2& 98.9 & 1.04 &  \textbf{0} &2,497 &1,170 &949 & Stanford Drone  \\ 
SMOT \cite{dicle2013way} &Faster RCNN    &36/24 & $1417\times 2019$ &99.1 &\textbf{100.0} &91.8 &91.9 &97.3 &1.38 &17,212 &38,846 &2,275 &3,926 & Stanford Drone\\  
SORT \cite{bewley2016simple} &Faster RCNN    &36/24 & $1417\times 2019$ &99.5 &98.1 &\textbf{95.7} &\textbf{96.0} &98.0 &1.05 &20 &32,436 &\textbf{957} &952 & Stanford Drone\\ 
SLSTM \cite{alahi2016social} &Faster RCNN    &36/24 & $1417\times 2019$ &99.3 &99.9 &89.6 &89.6 &99.8 &\textbf{0.13} &11 &841 &3, 630 &906 & Stanford Drone\\
\bottomrule[1.5pt]
\end{tabular}}
\label{tab:ExpMOT3}
\end{table*}

\begin{table*}[!t]
\caption{Performance comparisons of UAV exclusive tracking networks and classic tracking networks for the UAVDT dataset taken MOTA, MOTP, etc as evaluation indexes. The best performers are highlighted in bold.}
\small 
\centering
\resizebox{1\textwidth}{!}{
\begin{tabular}{l||l|c|c|c|c|c|c|c|c|c|c|c|c|c}
\toprule[1.5pt]
\textbf{Method}& \textbf{Framework} & \textbf{Train/Test(seq)} & \textbf{Image Size} &\textbf{MOTA} & \textbf{MOTP}& \textbf{IDF1} &\textbf{IDP} &\textbf{MT(\%)} &\textbf{ML(\%)} & \textbf{FP}  & \textbf{FN}& \textbf{IDS} & \textbf{FM} & \textbf{Exp. Data}\\ 
\hline \hline
\multirow{2}{*}{IPGAT\cite{yu2020conditional}} & LSTM+ CGAN   &\multirow{2}{*}{30/20} & \multirow{2}{*}{$1080\times 540$} &\multirow{2}{*}{39.0}&\multirow{2}{*}{72.2} &\multirow{2}{*}{49.4} &\multirow{2}{*}{63.2} &\multirow{2}{*}{37.4} &\multirow{2}{*}{25.2} &\multirow{2}{*}{42, 135} &\multirow{2}{*}{163, 837}&\multirow{2}{*}{2,091} &\multirow{2}{*}{10,057} & \multirow{2}{*}{UAVDT}\\ 
& +Siamese & &  & &   & &   & &  & &  &  & & \\ 
OSIM\cite{rs11182155} & YOLOv3 & - & $1080\times 540$ & \textbf{88.7} & - & - & - & - & - & \textbf{8} & \textbf{610} & - & - & UAVDT\\ 
Self-balance\cite{yu2019self} & LSTM & 30/20 & $1080\times 540$ & 38.6 & 72.1 & 48.5 & 61.1 & \textbf{38.9} & \textbf{24.4} & 44,724 & 160,950 & 3,489 & 11,796 & UAVDT\\ 
UAV\_MOT1 \cite{dike2021robust} &Faster RCNN    &30/20 & $1417\times 2019$ &40.3 &74.0 &\textbf{55.0} &\textbf{67.0} &-&-&30,065 &150,837 &\textbf{1,091} &\textbf{3,057} & UAVDT\\ 
RLSTM\cite{milan2016online}  &Faster RCNN &30/20 & $1080\times 540$ &25.6 &69.1 & 31.3 &38.6 &36.7 &25.7 &71,955 &180,461 &1,333 &13,088 & UAVDT\\  
SLSTM \cite{alahi2016social}&Faster RCNN &30/20 & $1080\times 540$ &37.9 &72.0&37.2 &46.8&38.2 &\textbf{24.4} &44,783 &161,009 &6,048 &12,051& UAVDT\\ 
SORT\cite{bewley2016simple}&Faster RCNN &30/20 & $1080\times 540$ &39.0 &\textbf{74.3}&43.7 &58.9&33.9 &  28.0 &33,037 &172,628 &2,350 &5,787& UAVDT\\ 
RMOT\cite{yoon2015bayesian} &Faster RCNN &30/20 & $1080\times 540$ & -39.8 &72.3&33.3 &27.8&36.7 & 25.7 &319,008 &151,485 &5,973 &5,897& UAVDT\\
SMOT\cite{dicle2013way}  &Faster RCNN &30/20 & $1080\times 540$ &33.9 & 72.2 & 45.0 &55.7 &36.7 &25.7 &57,112 &166,528 &1,752 &9,577 & UAVDT\\ 
CEM\cite{milan2013continuous}  &Faster RCNN    &30/20&$1080\times 540$ &-7.3 & 69.6 & 10.2 &19.4 &7.3 &68.6 & 72,378 & 290,962 &2,488 & 4,248 & UAVDT\\ 
GOG\cite{pirsiavash2011globally}  &Faster RCNN    &30/20&$1080\times 540$ &34.4 &72.2 & 18.0 &23.3 &35.5 &25.3 &41,126 &168,194 &14,301 &12,516 & UAVDT \\ 
IOUT\cite{bochinski2017high}&Faster RCNN   &30/20 &$1080\times 540$ &36.6 &72.1 &23.7 &30.3  &37.4 &25.0 &42,245 &163,881 &9,938 &10,463& UAVDT\\ 
\bottomrule[1.5pt]
\end{tabular}}
\label{tab:ExpMOT4}
\end{table*}

\subsection{Evaluation of Object Detection from UAV-borne Video}
For object detection from UAV-borne video, the common indicators to evaluate object detection methods are the same as UAV-borne image, including $AP^{IoU=0.50:0.05:0.95}$, $AP^{IoU =0.50}$,$AP^{IoU=0.75}$, $AR^{max=1}$, $AR^{max=10}$, $AR^{max=100}$ and $AR^{max =500}$. Table \ref{tab:ExpDet} lists the public quantitative results of some state-of-the-art and baseline detection works. Among them, four works are the UAV preserves object detection works, and the experimental results are mainly focused on the VisDrone dataset. TDFA with a two-stream refined flowNet (SPyNet) pipeline, which is robust to small-scale objects and can achieve the best performance among all comparison methods, i.e., 27.27\% AP score on the VisDrone-VID validation set. MPFPN with parallel branch is ranked as the second and third with 33.72\% and 29.05\% AP scores.

We have also summarized the results of 9 baseline methods in the VisDrone-VID challenge, including CFE-SSDv2 \cite{zhao2018cfenet}, FGFA$^*$ \cite{zhu2017flow}, RefineDet \cite{zhang2018single,zhu2018visdrone}, RetinaNet \cite{lin2017focal}, detection and tracking (D\&T) \cite{feichtenhofer2017detect}, FPN$^*$ \cite{lin2017feature}, CornerNet$^*$ \cite{law2018cornernet}, CenterNet$^*$ \cite{zhou2019objects}, and Faster R-CNN$^*$ \cite{ren2015faster}. The experiment results were instructed in accordance, with three non-overlapping subsets, 56 video sequences with 24,198 frames for the training set, 16 video sequences with 6,322 frames for testing, and the remaining sequences are for validation. Obviously, the detection performance of object detection in video yields better object detection in an image, and detection results amendment by context information plays a decisive role. Moreover, a small object is an inevitable problem of object detection in video. Therefore, CFE-SSD with small objects friendliness, FGFA assisting the current frame by adopting the front and back frames information, and D\&T with ROI tracking to associate adjacent frames, obtain a better detection performance.

Besides the VisDrone dataset, some other datasets are also used, such as Okutama-Action and UAVDT. Compared with five baseline works in Table \ref{tab:Okutama/UAVDTVid}, TDFA experimented Okutama-Action dataset has achieved the best detection performance, i.e., 87.18\% AP\_50 value on the Okutama-Action test dataset. STDnet-ST with Spatio-temporal ConvNet and STDnet experimented Okutama-Action dataset, have achieved 34.60\% AP for objects under $16\times 16$ pixels.

\begin{table*}[!t]
\centering
\caption{Computation Cost of Statistical Object Detection Approaches for UAV exclusive}
\begin{tabular}{l||l|l|l|c|c}
\toprule[1.5pt]
Reference & Network Pipeline & Image Size & Exp. environment &Times/fps & Year\\
\hline \hline
RRNet\cite{chen2019rrnet} &HourGlass &$512\times 512$ & - &- &2019\\ 
Wu et al\cite{wu2019DDCLS}&  YOLOv3 &$1080\times 640$& Workstation(NVIDIA
Tesla K80/-)&15&2019\\
SlimYOLOv3\cite{zhang2019slimyolov3}&YOLOv3-SPP3-90 & $832 \times 832$ &  Workstation(NVIDIA GTX 1080Ti/-)&28.3&2019\\
NDFT\cite{wu2019delving} & ResNet101 & - & Workstation(-/-)& -&2019\\
ClusDet\cite{ClusDet2019}& ResNeXt101& $2000\times 1500$ &Workstation(NVIDIA GTX 1080 Ti/-)&1.3 &2019\\ 
CenterNet\cite{pailla2019object}&  HourGlass104 & $1024\times 1024$ &-&-& 2019 \\ 
Yang et al\cite{yang2019effective}&  VGG16 & $320\times 320$& Worstation(NVIDIA GTX-1080Ti/12GB)&58 &2019\\ 
Zhang et al\cite{zhang2019dense} & ResNet50 & - &  Workstation(NVIDIA GeForce 1060/6GB) & -&2019\\ 
FS-SSD\cite{liang2019small} &VGG16 &$512\times 512$ &   Workstation(NVIDIA TITAN X (Pascal)/12GB)
& 18.3& 2019\\
SAMFR\cite{wang2019spatial}& DetNet59&$419\times 419$ & -& 10 &2019\\
MSOA-Net\cite{MSOA-Net2020}&  ResNet50 & $1333\times 800$ & Workstation(NVIDIA TITAN-Xp/-) & & 2020\\
GDF-Net\cite{GDF-Net2020}&  ResNet50 & $1200\times 675$ & Workstation(NVIDIA Geforce RTX 2080ti/11GB) &17.9 &2020\\ 
HRDNet\cite{liu2020HRDNet}&ResNeXt50+101 & $960\times 1360$ & Workstation(NVIDIA GTX 2080Ti/-) &0.7 &2020 \\ 
D-A-FS SSD\cite{ICIT2020}& -& -& - &-  & 	2020\\ 
UAV-YOLO\cite{liu2020uav}& YOLOv3 &  $608\times 608$ & Workstation(NVIDIA GTX Titan XP/64GB) & 20&2020 \\
SyNet\cite{albaba2020synet} & CenterNet &   -
&- & -&2020\\ 
ComNet\cite{li2020comnet}& YOLOv3 &   $416\times 416$ &Workstation(NVIDIA GTX 1080Ti/12GB)& 20&2020 \\ComNet\cite{li2020comnet}& YOLOv3 &   $416\times 416$& laptop(Intel Core i5-8300/4GB)&3 &2020\\ComNet\cite{li2020comnet}& YOLOv3 &   $416\times 416$&   Jason Nano(Tegra X1/4GB)&2& 2020\\
MPFPN\cite{liu2020small}&  ResNets101 &   $1440\times 800$
& Workstation(NVIDIA GTX 1080Ti/-) & 2.1 &2020\\ 
D2Det\cite{cao2020d2det}&  ResNet101 &   $1333\times 800$ & Workstation(NVIDIA GTX Titan Xp/-) &5.9 & 2020\\
DAGN\cite{zhang2019dagn}& YOLOv3 &   $512\times 512$&Workstation(NVIDIA GeForce GTX 1080Ti/11GB)&25.1 &2020\\
GANet\cite{yuanqiang2020guided}& VGG-16/ResNet50&  $512\times 512$&Workstation(NVIDIA GTX Titan Xp/-)&-&2020\\
DAN\cite{jadhav2020aerial}& Resnet-50&$1500\times 1000$ & -& - &2020\\
Zhang et al\cite{zhang2020coarse}& PeleeNet & $304\times 304$& Workstation(NVIDIA
TITAN X (Pascal)/12GB)
& 23.6 & 2020\\
DSHNet\cite{yu2021towards}&  ResNet50& $2000\times 1500$ & Workstation(NVIDIA 1080Ti/11GB) & 10.8& 2021\\ 
DNOD\cite{TIAN2021292}&VGG19+CSPDarknet53 & $608\times 608$ &  Workstation(NVIDIA GeForce RTX 2080ti/6GB)&38.3&2021\\
ECas\_RCNN\cite{ECascade-RCNN}&  ResNet50 & $1450\times 800$ & Workstation(NVIDIA RTX
2080Ti/-) &- & 2021\\ 
DSYolov3\cite{DSYolov32021}&  Yolov3 & $416\times 416$& Workstation(NVIDIA GTX 1080Ti/-) &13.7 &2021\\ 
Cas\_RCNN+FPN\cite{youssef2021automatic}&ResNet101 & $1500 \times 2000$ &  -&-&2021\\
\bottomrule[1.5pt]
\end{tabular}
\label{tab:SODCost}
\end{table*}

\begin{table*}[!t]
\centering
\caption{Computation Cost of DL-based Video Object Detection Approaches for UAV exclusive}
\resizebox{1\textwidth}{!}{
\begin{tabular}{l||l|l|l|c|c}
\toprule[1.5pt]
Reference & Network Pipeline & Image Size & Exp. environment &Times/fps & Year\\
\hline \hline
STCA\cite{Pi_2019_ICCV} &SSD+FCOS+SiamFC &$300\times 300$ & Workstation(NVIDIA GeForce GTX 1080/-) &- &2019\\ 
Abughalieh et al\cite{abughalieh2019video}& FAST & $320\times 240$& laptop(Core i7-2670QM/6GB)&26.3&2019\\Abughalieh et al\cite{abughalieh2019video}& FAST &$320\times 240$ & Embedded(Raspberry Pi 2/1GB)&10.8&2019\\
Nousi et al\cite{nousi2019embedded}& Tity YOLO   &$288\times 288$ & Embedded(Robot Operating System) & 23 &2019 \\
SCNN\cite{wang2019scnn} &ResNet34&$224\times 224$& Workstation(NVIDIA GeForce GTX 1080/-)&246 &2019 \\ 
Zhang et al\cite{zhang2020drone} & Cas\_R-CNN+IRR-PWC \cite{sun2018pwc} &$720\times 1280$&-&-&2020\\
MOR-UAVNet\cite{mandal2020mor}& MOR-UAVNetv1\-4 & $608\times 608$& Workstation(NVIDIA RTX 2080 Ti/11GB) &10.5&2020\\
TDFA\cite{xie2021two}& FlowNet+ Fea\_Aggregation &$720\times 1280$ & Workstation(NVIDIA GeForce GTX TITAN X/12GB)  &3.8& 2021 \\
STDnet-ST\cite{bosquet2021stdnet}& STDnet+ConvNet& $1280\times 720$ &- &-& 2021\\ STDnet-ST\cite{bosquet2021stdnet}&STDnet+ConvNet& $1920\times 1080$ & -&-& 2021\\ STDnet-ST\cite{bosquet2021stdnet}&STDnet+ConvNet& $1024\times 540$ & -&- &2021\\ 
\bottomrule[1.5pt]
\end{tabular}}
\label{tab:VIDCost}
\end{table*}

\subsection{Evaluation of Object Tracking from UAV-borne Video}
For object tracking from UAV-borne video, the common way to evaluate object detection methods, including multiple object tracking accuracy (MOTA), multiple object tracking precision (MOTP), identification precision (IDP), identification F1 score (IDF1), false alarms per frame (FAF), the number of mostly tracked targets (MT, more than 80\% of trajectories being covered by the ground truth), the number of mostly lost targets (ML, less than 20\% of trajectories being covered by the ground truth), the number of false positives (FP), the number of false negatives (FN), the number of ID switches (IDS), and the number of times a trajectory is Fragmented (FM). 

The IDF1 score is defined as 
\begin{equation}
IDF1 = \frac{2IDTP}{2IDTP+IDFP+IDFN},
\end{equation}
where IDTP is the number of true positive IDs, IDFP is the number of false positive IDs, and IDFN is the number of false negative IDs. In addition, some literature have still adopted the detection evaluation metrics, including $AP^{IoU=0.50:0.05:0.95}$, $AP^{IoU=0.25}$, $AP^{IoU =0.50}$,
$AP^{IoU=0.75}$.

Tables \ref{tab:ExpMOT1}-\ref{tab:ExpMOT4} summarize the quantitative comparison of several multiple object tracking methods on the challenging public UAV dataset. In Table \ref{tab:ExpMOT1}, the average rank of 10 metrics (i.e., MOTA, MOTP, IDF1, FAF, MT, ML, FP, FN, IDS, and FM) is used to rank these approaches. TrackletNet Tracker (TNT), wins the VisDrone-MOT challenge dataset by the highest MOTA, IDF1, FP, IDS. We also report the accuracy of the trackers in AP as well as different object categories, including AP\_car, AP\_bus, AP\_trk, AP\_ped and AP\_van in Table \ref{tab:ExpMOT2}. The PAS tracker followed by the tracking-by-detection paradigm achieves the best performance, i.e., 50.80\% AP score on the VisDrone-MOT testing set. HMTT based on SOT achieves a 28.67\% AP score on the VisDrone-MOT validation set. With the exception of VisDrone dataset, IPGAT achieves the best tracking performance for the UAVDT and Stanford Drone testing dataset in Tables \ref{tab:ExpMOT3} and \ref{tab:ExpMOT4}, in terms of IDF1, MT, ML, and FN, by estimating object motion and UAV movement as individual and global motions, respectively.

The rest of Tables \ref{tab:ExpMOT1} to \ref{tab:ExpMOT4} are the results from baseline methods for the three MOT datasets. The results are based on Faster RCNN detection input for convenient comparison. For VisDrone-MOT dataset consists of 79 video sequences in total, including 56 video sequences for the training set, 16 video sequences for testing, and the remainder is for validation. Under these settings, Ctrack with recovering long-time disappearance objects in the crowded scenes achieves the best tracking performance among all methods in the VisDrone testing dataset, in terms of the IDF1, MT, ML, and FN. For the UAVDT dataset under 50 sequences recorded in the traffic scenario from UAVs, 60\% for training and 40\% for testing. From Table \ref{tab:ExpMOT4}, it can be seen that SORT is superior on most metrics. Although these results are far from the requirements of practical application, they can provide feasible direction (e.g., the association of moving objects) and a reliable theoretical basis for future research. For the Stanford Drone dataset, the performance gap of the listed baseline method is minimal, maybe IOUT slightly better.

\subsection{Estimation of Computation Cost}
In this survey, all the reviewed methods have their own experimental environment, experimental data, and even source code. Considering the computation cost is directly related to speed, GPU, and backbone model, we list these three indexes of the UAV exclusive methods for the above three topics in Tables \ref{tab:SODCost}- \ref{tab:MOTCost}. Depending on the computing power of NVIDIA's GPU \footnote{\url{https://www.pianshen.com/article/13711825712/}}, the computation cost can be estimated with backbone network in the corresponding method.

\begin{table*}[!t]
\centering
\caption{Computation Cost of DL-based Multiple Object Tracking for UAV exclusive.}
\resizebox{1\textwidth}{!}{
\begin{tabular}{l||l|l|l|c|c}
\toprule[1.5pt]
Reference & Network Pipeline & Image Size & Exp. environment &Times/fps & Year\\
\hline \hline
Deep SORT\cite{wojke2017simple} &   Faster R-CNN+ Sort & $1920\times 1080$&Workstation(NVIDIA GeForce GTX 1050/-)&-&2017\\ 
SCTrack\cite{al2018multi}&  Faster R-CNN+YOLOv3 & -  &-&-&2018\\
TNT\cite{zhang2019eye}&  Faster-RCNN+ SVO+MVS+3d loc & -&-& -&2019\\
Zhou et al\cite{zhou2019uav}&Faster RCNN+ SPHP &$1080\times 540$ &Workstation(NVIDIA RTX 2080Ti/-) &-& 2019\\ 
\multirow{2}{*}{OSIM\cite{rs11182155}}& YOLOv3+Kalman filtering&\multirow{2}{*}{$2720\times 1530$} &Workstation(NVIDIA GeForce GTX 1080 Ti/-)&\multirow{2}{*}{30}&\multirow{2}{*}{2019}\\& +deep appearance feature & & Workstation(Intel UHD Graphics 630/-)&&\\
Self-balance\cite{yu2019self}& LSTM  &  $1080\times 540$ & Workstation(NVIDIA Titan X/32GB) &-&2019\\
Flow-tracker\cite{li2019multiple}&   Optical Flownet+IOU  &- & Workstation(NVIDIA GTX 1080Ti/-) &5&2019\\ 
HMTT\cite{pan2019multi}&  CenterNet+IOU+OSNet  & - &-&-&2019\\ 
Yang et al\cite{yang2019multiple}&  YOLOv3 + dense-trajectory-Voting   &  $1920\times 1080$ &  Workstation(NVIDIA GeForce GTX1080Ti/6GB)&8.6&2019\\ 
GGD\cite{ardo2019multi} &  Faster RCNN+GGD & -  & Workstation(NVIDIA GTX 1080/-) &-&2019\\
COMET\cite{marvasti2020comet}&  ResNet-50+Two-stream network  & $1080\times 540$  &  Workstation(NVIDIA Tesla V100/16GB) &24 &2019 \\
Abughalieh et al\cite{abughalieh2019video}&FAST & $320\times 240$& Laptop(Core i7-2670QM/6GB)&26.3&2019\\
Abughalieh et al\cite{abughalieh2019video}&FAST  &$320\times 240$ & Embedded(Raspberry Pi 2/1GB)&10.8&2019\\
IPGAT\cite{yu2020conditional}& SiameseNet+LSTM+CGAN &- & Workstation(NVIDIA Titan X/32GB) &-&2020\\  
Kapania et al\cite{kapania2020multi}&   YOLOv3+RetinaNet & - &-&-&2020 \\
PAS tracker\cite{stadler2020pas} &  Cascade R-CNN+Similarity   &-&-&-& 2020 \\
DAN\cite{jadhav2020aerial}& RetinaNet+DeepSORT  &  $1500\times 1000$  &  - & &2020\\
DQN\cite{dike2021robust}& Faster R-CNN &-&-&- &2021\\
Youssef et al\cite{youssef2021automatic}&Cascade R-CNN+FPN&$2000\times 1500$&Workstation(NVIDIA Quadro RTX5000/16GB)  &-&2021\\
\multirow{2}{*}{HDHNet\cite{huang2021multiple}}&   \multirow{2}{*}{HDHNet+DeepSORT+Cas\_RCNN} &$1280\times 720$ & \multirow{2}{*}{Workstation(NVIDIA TITAN RTX/24GB)} &4.3 &\multirow{2}{*}{2021}\\ &   &$2880\times 1620$ &  &2.1 &\\ 
\bottomrule[1.5pt]
\end{tabular}}
\label{tab:MOTCost}
\end{table*}
\section{Discussion and Conclusion} \label{sec:Con}
In this paper, deep learning approached in object and tracking of the remote sensing field has been systematically analyzed according to three UAV topics, i.e., SOD, VID, and MOT. The conclusions were drawn as follows.

\textbf{UAV data}: The public UAV-borne datasets for object detection and tracking are mainly visible data, and the largest image size is $3840\times 2160$ (VisDrone dataset). There are only one multiple source data called Vehicle dataset with visible-thermal infrared cameras equipped with drones. In label terms, the bounding box is not limited to horizontal bounding box strongly dependent on robustness to direction, even have oriented bounding box, e.g., in the Vehicle Dataset.

\textbf{DL Method}: This survey reviews DL-based object detection and tracking methods for UAV acquired data from three topics. In general, most classical DL methods, by appending extra modules available to UAV challenges, can be applied to these three topics. Specifically, considering different requirements for precision and speed, the existing static object detection methods especially for UAV are mainly based on YOLO (e.g., UAV-YOLO, ComNet, SlimYOLOv3, DAGN, etc), Faster RCNN (e.g., Dshnet, NDFT, D2det, etc), and SSD (e.g., FS SSD). Among them, YOLO, and SSD based methods are advantageous in speed. For VID and MOT, there are few methods especially designed for UAV data. Most literature are still about classical methods for natural scene data, such as Flownet, LSTM for VID, and DeepSort, SiamRPN for MOT. As a consequence, their performance is far from perfect, e.g., the highest AP for VisDrone-VID is just 65.2\%, and for VisDrone-MOT is just 50.80\%. Further effort is needed to solve the interaction of ground objects and the complexity of the tracking scenes. As for systems, the existing UAV object detection and tracking systems are mainly based on the classic DL method, where the speed can be guaranteed but the accuracy still needs to be improved.

\textbf{Computer platforms}: The image/video acquired by UAV in this review mainly belongs to the remote sensing community. In this community, DL-based methods are mainly carried out on various NVIDIA series GPU, e.g., TITAN Xp, RTX 2080Ti, GTX 1080Ti, etc. Their processing speed is roughly within the range of $0.2fps\sim 50fps$ with different image size. Although the research reviewed in \cite{abughalieh2019video,nousi2019embedded} designs object detection and tracking system using a Raspberry Pi 2 minicomputer process for object detection with 11fps, or using a Jetson TX2 embedded platform for object detection with 8.5fps, even 4.5fps \cite{deng2020lightweight} and object tracking with 15fps, but lacks generality.

Object detection with tracking reflects a perfect union in engineering practice. With tracking assistance, detection becomes stable and exhibits no jitter. Meanwhile, fine labels and ID information of objects with the same class are also given. Through automatic analysis and extraction of trajectory features, false and missed detection rates can be significantly reduced. In the near future development of object detection and tracking in UAV remote sensing is expected and new techniques will emerge to improve these metrics even further. In addition, efficiently processing massive multi-source UAV remote sensing data are worth consideration. UAVs equipped with different sensors, e.g., visible, infrared, thermal infrared, multispectral, hyperspectral sensors, can integrate a variety of sensing modalities to make use of their complementary properties, which further realizes more robust and accurate object tracking and detection.

\bibliographystyle{ieeetr}
\bibliography{reference}

\begin{IEEEbiography}[{\includegraphics[width=1in,height=1.25in,clip,keepaspectratio]{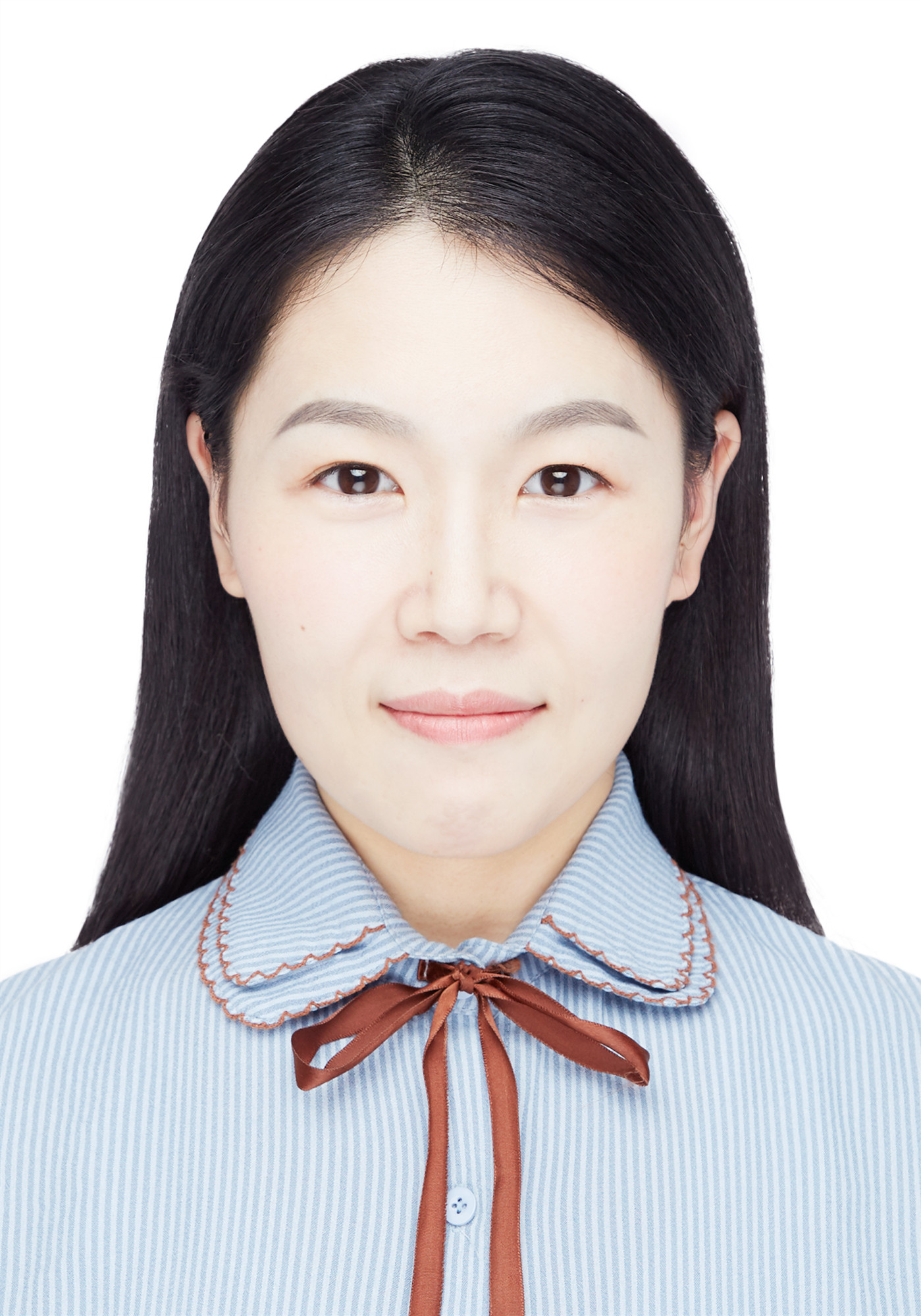}}]{Xin Wu}
(S'19--M'20) received the M.Sc. degree in Computer Science and Technology from the College of Information Engineering, Qingdao University, Qingdao, China, in 2014, the Ph.D. degree from the School of Information and Electronics, Beijing Institute of Technology (BIT), Beijing, China, in 2020.

In 2018, she was a visiting student at the Photogrammetry and Image Analysis department of the Remote Sensing Technology Institute (IMF), German Aerospace Center (DLR), Oberpfaffenhofen, Germany. She is currently a Postdoctoral researcher in the School of Information and Electronics, BIT, Beijing, China. Her research interests include signal / image processing, fractional Fourier transform, deep learning and their applications in biometrics and geospatial object detection.

She was a recipient of the Jose Bioucas Dias award for recognizing the outstanding paper at the Workshop on Hyperspectral Imaging and Signal Processing: Evolution in Remote Sensing (WHISPERS) in 2021.
\end{IEEEbiography}

\vskip -2\baselineskip plus -1fil

\begin{IEEEbiography}[{\includegraphics[width=1in,height=1.25in,clip,keepaspectratio]{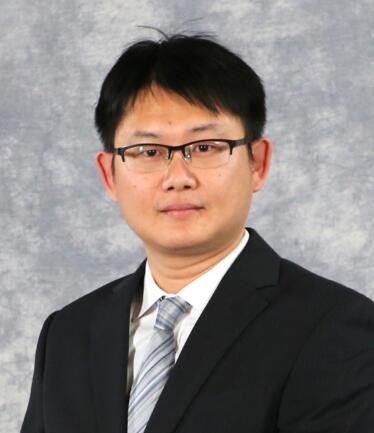}}]{Wei Li}(S'11--M'13--SM'16) received the B.E.degree in telecommunications engineering from Xidian University, Xi'an, China, in 2007, the M.S. degree in information science and technology from Sun Yat-Sen University, Guangzhou, China, in 2009, and the Ph.D. degree in electrical and computer engineering from Mississippi State University, Starkville, MS, USA, in 2012.

Subsequently, he spent 1 year as a Postdoctoral Researcher at the University of California, Davis, CA, USA. He is currently a professor with the School of Information and Electronics, Beijing Institute of Technology. His research interests include hyperspectral image analysis, pattern recognition, and data compression.

He is currently serving as Associate Editor for the IEEE Transactions on Geoscience and Remote Sensing (TGRS), IEEE Journal of Selected Topics in Applied Earth Observations and Remote Sensing (JSTARS), and IEEE Signal Processing Letters (SPL). He has published more than 150 peer-reviewed articles and 100 conference papers totally cited by 7500 times (Google Scholar). He received the JSTARS Best Reviewer in 2016 and TGRS Best Reviewer award in 2020 from IEEE Geoscience and Remote Sensing Society (GRSS), and the Outstanding Paper award at IEEE International Workshop on Hyperspectral Image and Signal Processing: Evolution in Remote Sensing (Whispers), 2019.
\end{IEEEbiography}

\vskip -2\baselineskip plus -1fil

\begin{IEEEbiography}[{\includegraphics[width=1in,height=1.25in,clip,keepaspectratio]{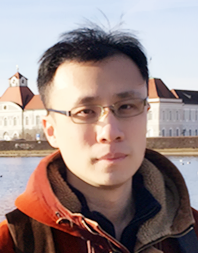}}]{Danfeng Hong}
(S'16--M'19--SM'21) received the M.Sc. degree (summa cum laude) in computer vision from the College of Information Engineering, Qingdao University, Qingdao, China, in 2015, the Dr. -Ing degree (summa cum laude) from the Signal Processing in Earth Observation (SiPEO), Technical University of Munich (TUM), Munich, Germany, in 2019. 

Since 2015, he has been a Research Associate at the Remote Sensing Technology Institute (IMF), German Aerospace Center (DLR), Oberpfaffenhofen, Germany. He is currently a Research Scientist and leads a Spectral Vision Working Group at IMF, DLR. He is also an Adjunct Scientist at GIPSA-lab, Grenoble INP, CNRS, Univ. Grenoble Alpes, Grenoble, France. His research interests include signal / image processing and analysis, hyperspectral remote sensing, machine / deep learning, artificial intelligence, and their applications in Earth Vision.

Dr. Hong is an Editorial Board Member of Remote Sensing and a Topical Associate Editor of the IEEE Transactions on Geoscience and Remote Sensing (TGRS). He was a recipient of the Best Reviewer Award of the IEEE TGRS in 2021 and the Jose Bioucas Dias award for recognizing the outstanding paper at the Workshop on Hyperspectral Imaging and Signal Processing: Evolution in Remote Sensing (WHISPERS) in 2021. He is also a Leading Guest Editor of the International Journal of Applied Earth Observation and Geoinformation, the IEEE Journal of Selected Topics in Applied Earth Observations, and Remote Sensing.
\end{IEEEbiography}

\vskip -2\baselineskip plus -1fil

\begin{IEEEbiography}[{\includegraphics[width=1in,height=1.25in,clip,keepaspectratio]{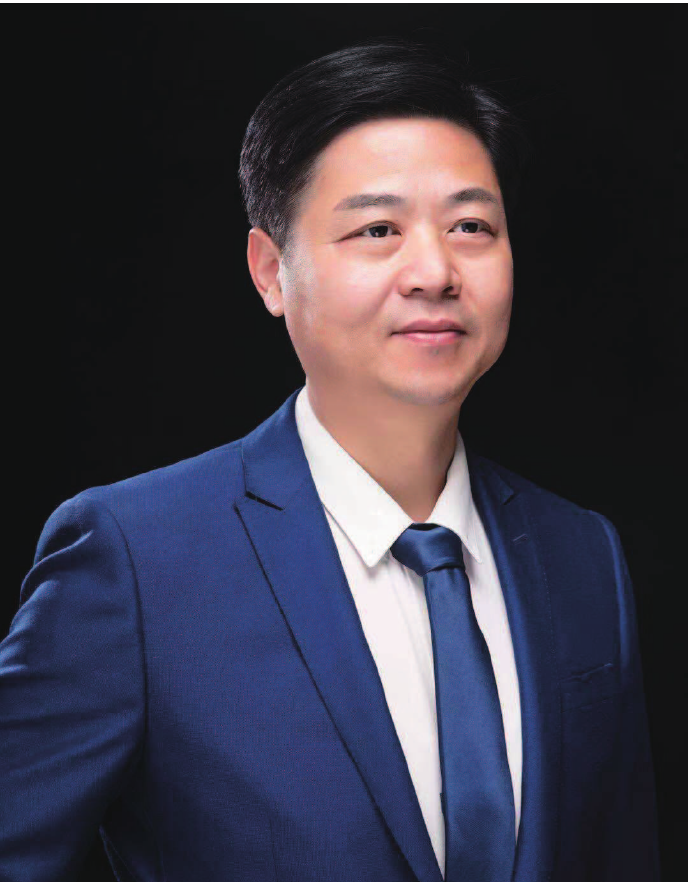}}]
{Ran Tao}(M'00–-SM'04) received the B.S. degree from the Electronic Engineering Institute of PLA, Hefei, China, in 1985, and the M.S. and Ph.D. degrees from the Harbin Institute of Technology, Harbin, China, in 1990 and 1993, respectively. In 2001, he was a Senior Visiting Scholar with the University of Michigan, Ann Arbor, MI, USA. 

He is currently a Professor with the School of Information and Electronics, Beijing Institute of Technology, Beijing, China. He has authored 3 books and more than 180 peer-reviewed journal articles. His current research interests include fractional signal and information processing with applications.

Dr. Tao was the recipient of the National Science Foundation of China for Distinguished Young Scholars in 2006, and the First Prize of Science and Technology Progress in 2006 and 2007, and the First Prize of Natural Science in 2013, both awarded by the Ministry of Education. He was a Distinguished Professor of the Changjiang Scholars Program in 2009. He was a Chief Professor of the Program for Changjiang Scholars and Innovative Research Team in University from 2010 to 2012. He has been a Chief Professor of the Creative Research Groups of the National Natural Science Foundation of China since 2014. He has been awarded the Famous Teachers of Higher Education in Beijing in 2018. He is currently the Vice Chair of the International Union of Radio Science China Council.
\end{IEEEbiography}

\vskip -2\baselineskip plus -1fil

\begin{IEEEbiography}[{\includegraphics[width=1in,height=1.25in,clip,keepaspectratio]{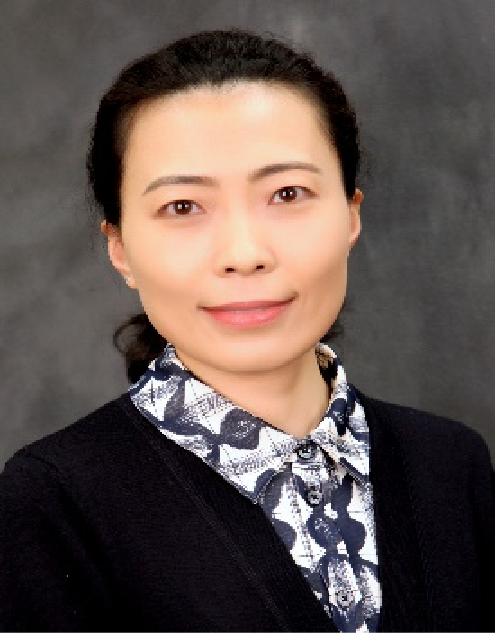}}]{Qian Du} (M'00--SM'05--F'18) received the Ph.D. degree in electrical engineering from the University of Maryland at Baltimore County, Baltimore, MD, USA, in 2000. She is currently the Bobby Shackouls Professor with the Department of Electrical and Computer Engineering, Mississippi State University, MS, USA. Her research interests include hyperspectral remote sensing image analysis and applications, pattern classification, data compression, and neural networks.

Dr. Du is a fellow of the SPIE-International Society for Optics and Photonics. She received the 2010 Best Reviewer Award from the IEEE Geoscience and Remote Sensing Society. She was the Co-Chair of the Data Fusion Technical Committee of the IEEE Geoscience and Remote Sensing Society from 2009 to 2013, and the Chair of the Remote Sensing and Mapping Technical Committee of the International Association for Pattern Recognition from 2010 to 2014. She has served as an Associate Editor of the IEEE JOURNAL OF SELECTED TOPICS IN APPLIED EARTH OBSERVATIONS AND REMOTE SENSING, the Journal of Applied Remote Sensing, and the IEEE SIGNAL PROCESSING LETTERS. She is the Editor-in-Chief of the IEEE JOURNAL OF SELECTED TOPICS IN APPLIED EARTH OBSERVATIONS AND REMOTE SENSING from 2016 to 2020. 

\end{IEEEbiography}


\end{document}